\documentclass[10pt,twocolumn,letterpaper]{article}

\usepackage{iccv}
\usepackage{times}
\usepackage{epsfig}
\usepackage{graphicx}
\usepackage{amsmath}
\usepackage{amssymb}

\usepackage{booktabs}
\usepackage{multirow}
\usepackage{tabularx}
\usepackage{enumerate} 
\usepackage{xcolor,colortbl}
\usepackage{enumitem}
\usepackage{bm}
\usepackage{algorithmic}
\usepackage{algorithm}
\usepackage{caption}
\usepackage{subcaption}
\usepackage{pifont}
\usepackage{ulem}
\usepackage[titletoc]{appendix}

\usepackage[breaklinks=true,bookmarks=false]{hyperref}
\usepackage[noabbrev]{cleveref}

\iccvfinalcopy 

\def\iccvPaperID{2395}

\ificcvfinal\fi

\def\c{{\boldsymbol c}}
\def\x{{\boldsymbol x}}
\def\y{{\boldsymbol y}}
\def\z{{\boldsymbol z}}

\def\W{{\boldsymbol W}}
\def\D{{\boldsymbol D}}
\def\Y{{\boldsymbol Y}}

\newcommand{\xmark}{\ding{55}}%

\DeclareMathOperator*{\argmax}{arg\,max}

\definecolor{Gray}{gray}{0.9}
\definecolor{LightCyan}{rgb}{0.88,1,1}

\newcolumntype{a}{>{\columncolor{Gray}}c}
\newcolumntype{b}{>{\columncolor{white}}c}
\def\algo{{\textsc{NGC}}}

\begin{document}

%%%%%%%%% TITLE
\title{NGC: A Unified Framework for Learning with Open-World Noisy Data}

\author{Zhi-Fan Wu$^{1,2}$\thanks{Equal contribution. $^\dagger$Corresponding author. This work was supported by Alibaba Group through Alibaba Innovative Research Program and the National Natural Science Foundation of China (61772262).}, Tong Wei$^{1}$\footnotemark[1], Jianwen Jiang$^{2}$\footnotemark[1], Chaojie Mao$^2$, Mingqian Tang$^2$, Yu-Feng Li$^{1\dagger}$\\
$^1$State Key Laboratory for Novel Software Technology, Nanjing University, Nanjing, China\\
$^2$Alibaba Group, China\\
{\tt\small \{wuzf, weit\}@lamda.nju.edu.cn, liyf@nju.edu.cn} \\
{\tt\small \{jianwen.jjw, chaojie.mcj, mingqian.tmq\}@alibaba-inc.com}
}

\maketitle
\ificcvfinal\thispagestyle{empty}\fi

%%%%%%%%% ABSTRACT

\begin{abstract}
The existence of noisy data is prevalent in both the training and testing phases of machine learning systems, which inevitably leads to the degradation of model performance. There have been plenty of works concentrated on learning with in-distribution (IND) noisy labels in the last decade, \textit{i.e.}, some training samples are assigned incorrect labels that do not correspond to their true classes. Nonetheless, in real application scenarios, it is necessary to consider the influence of out-of-distribution (OOD) samples, \textit{i.e.}, samples that do not belong to any known classes, which has not been sufficiently explored yet. To remedy this, we study a new problem setup, namely Learning with Open-world Noisy Data (LOND). The goal of LOND is to simultaneously learn a classifier and an OOD detector from datasets with mixed IND and OOD noise. In this paper, we propose a new graph-based framework, namely Noisy Graph Cleaning (\algo), which collects clean samples by leveraging geometric structure of data and model predictive confidence. Without any additional training effort,~\algo~can detect and reject the OOD samples based on the learned class prototypes directly in testing phase. We conduct experiments on multiple benchmarks with different types of noise and the results demonstrate the superior performance of our method against state of the arts.
\end{abstract}

%%%%%%%%% BODY TEXT
\section{Introduction}
Deep neural networks (DNNs) have gained popularity in a variety of applications. Despite their success, DNNs often rely on the availability of large-scale labeled training datasets. In practice, data annotation inevitably introduces label noise, and it is extremely expensive and time-consuming to clean up the corrupted labels. The existence of label noise can be problematic for overparameterized deep networks, as they may overfit to label noise even on randomly-assigned labels~\cite{zhang2017understanding}. Therefore, mitigating the effects of noisy labels becomes a critical issue.

\begin{figure}[t]
    \centering
    \includegraphics[width=0.95\linewidth]{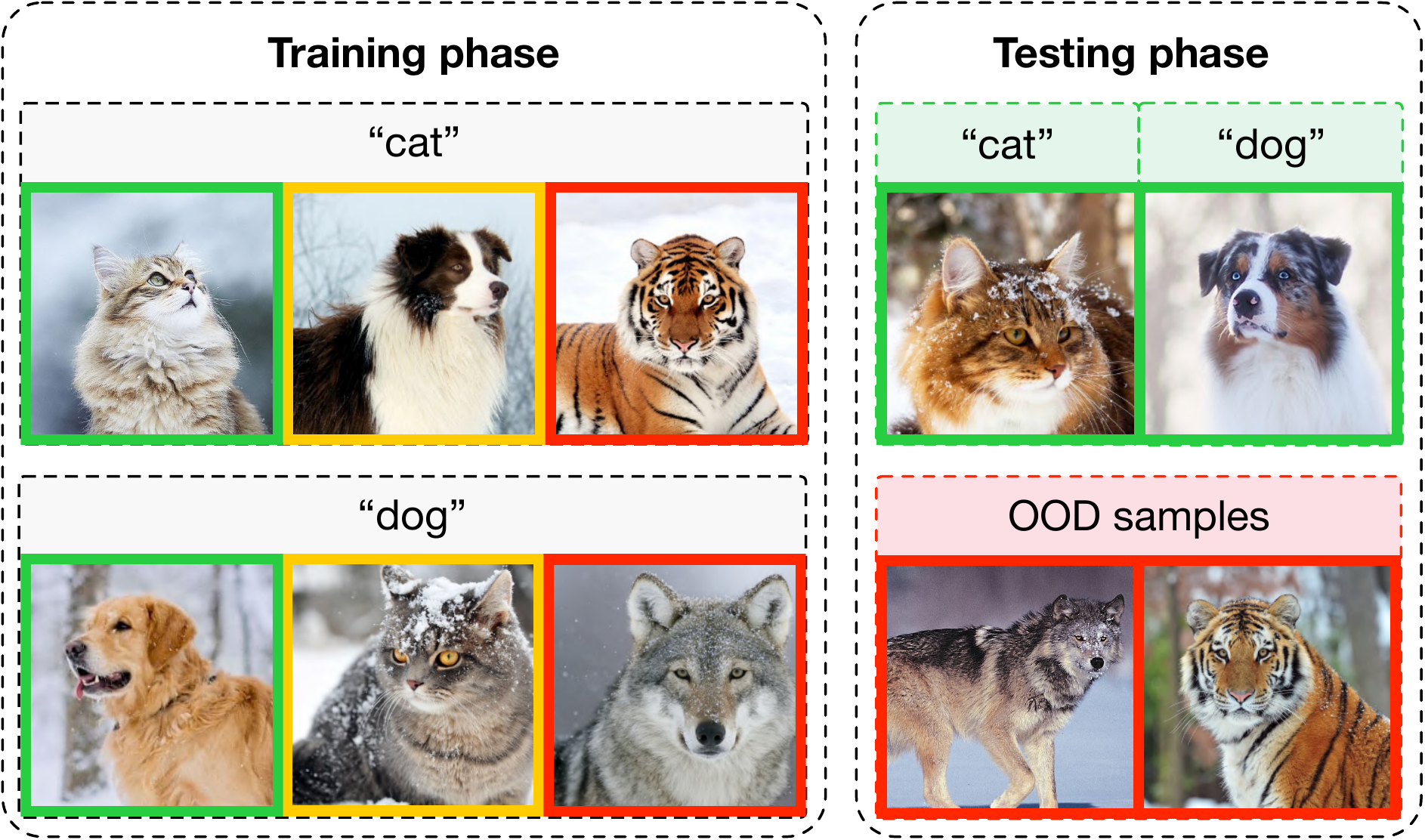}
    \caption{A demonstration of the LOND setup. We use green boxes to represent clean samples while yellow and red boxes are IND and OOD noisy samples, respectively.}
    \label{fig:problem_illustration}
\end{figure}

When learning with noisy labels (LNL), plenty of promising methods have been proposed to improve the generalization~\cite{Tanaka2018joint,guo2018curriculumNet,wang2019symmetric,li2019learningtolearn,wang2019co-mining,zhang2020distilling,xia2020part,yang2020webly,xia2021robust}. Many existing methods work by analyzing output predictions to identify mislabeled samples~\cite{yi2019probabilistic,pleiss2020identifying,li2020dividemix} or reweighting samples to alleviate the influence of noisy labels~\cite{ren2018learning,arazo2019unsupervised}. Note that, these methods are particularly designed to deal with in-distribution (IND) label noise. Some other works also consider the existence of out-of-distribution (OOD) noise in training datasets~\cite{wang2018iterative,lee2019rog}. Their basic assumption is that clean samples are clustered together while OOD samples are widely scattered in the feature space.

\begin{figure}[t]
    \centering
    \includegraphics[width=1\linewidth]{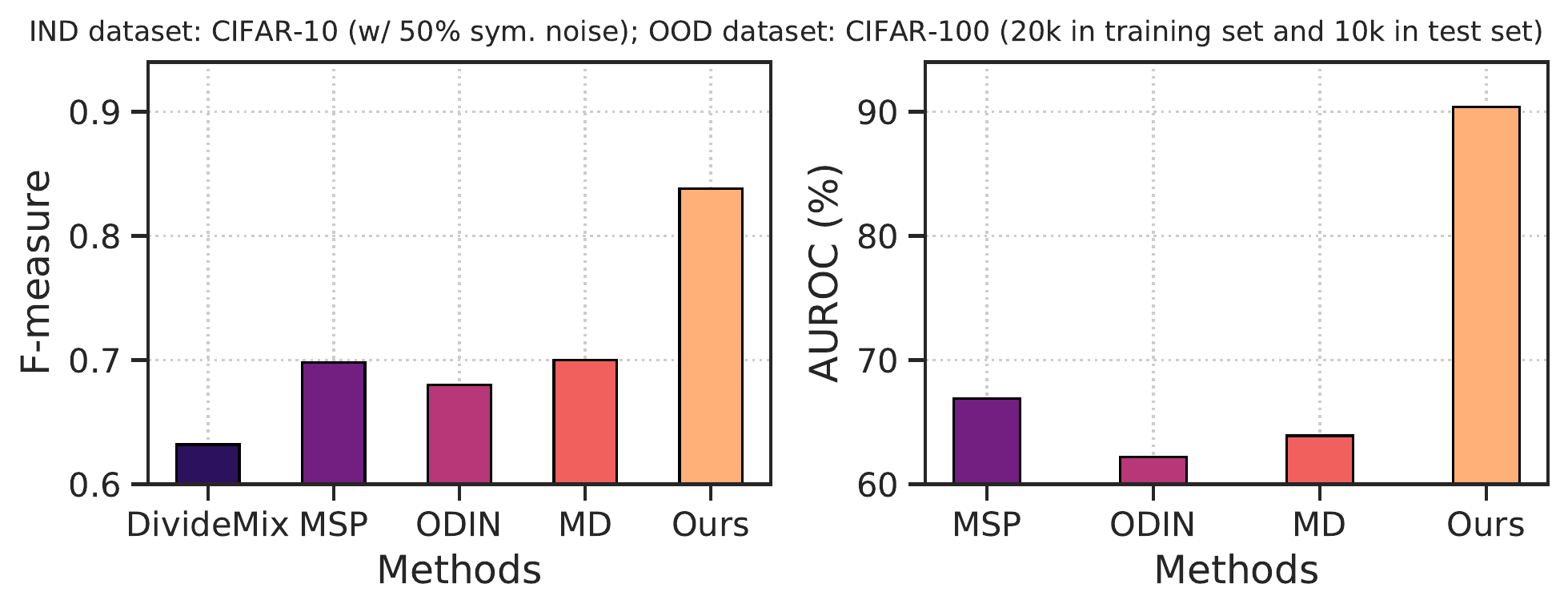}
    \caption{Performance on testing dataset with extra OOD samples. MSP, ODIN, MD are combined with DivideMix.}
    \label{fig:fmeasure-auroc-demo}
\end{figure}

Although significant performance improvement is achieved, most existing LNL works only take account of OOD samples in training phase, while the existence of OOD samples in testing phase is neglected, which is crucial for machine learning systems in real applications~\cite{hendrycks2017msp,lee2018mahalanobis,sehwag2021ssd}. In this paper, we study this practical problem, i.e., the existence of both IND and OOD noise in training phase, as well as the presence of OOD samples in testing phase. We name this new setup as learning with open-world noisy data (LOND). An illustration of the LOND setup can be found in Figure~\ref{fig:problem_illustration}. A straightforward approach to address LOND is to combine LNL methods with OOD detectors~\cite{hendrycks2017msp,liang2018odin,lee2018mahalanobis}. However, we empirically find that such direct combinations lead to unsatisfactory results as shown in Figure~\ref{fig:fmeasure-auroc-demo}. Therefore, obtaining models that can handle IND and OOD noise in both training and testing phases remains challenging.

To address the LOND problem, we present Noisy Graph Cleaning (\algo), a unified framework for learning with open-world noisy data. Different from previous LNL methods that utilize either model predictions~\cite{li2020dividemix,pleiss2020identifying,li2021learning,li2021mopro} or neighborhood information~\cite{wang2018iterative,wu2020topofilter}, where the interaction between model predictions and geometric structure of data is neglected,~\algo~simultaneously takes advantage of output confidence and the geometric structure. With the help of graph structure, we find that the confidence-based strategy can break the connectivity between clean and noisy samples, which significantly facilitates the geometry-based strategy. In specific,~\algo~iteratively constructs the nearest neighbor graph using latent representations of training samples. Given the graph structure,~\algo~corrects IND noisy labels by aggregating information from neighborhoods through soft pseudo-label propagation. Then, to remove the OOD and remaining obstinate IND noise, we present subgraph selection. It first degrades the connectivity between clean and noisy samples by removing samples with low-confidence predictions. Then, subgraphs corresponding to the largest connected component are constructed for each class. Moreover,~\algo~employs the devised contrastive losses~\cite{wu2018unsupervised,chen2020simclr,khosla2020supervised} to refine the representations from both instance-level and subgraph-level, which in return benefits label correction and subgraph selection. At test time,~\algo~can readily detect and reject OOD samples by calculating distances to learned class prototypes.

The main contributions of this work are:
\begin{enumerate}[topsep=3pt]
\item We study a new problem, that is, the training set contains both IND and OOD noise and the test set contains OOD samples, which is practical in real applications. 
\item We propose a new graph-based noisy label learning framework,~\algo, which corrects IND noisy labels and sieves out OOD samples by utilizing the confidence of model predictions and geometric structure of data. Without any additional training effort,~\algo~can detect and reject OOD samples at testing time.

\item We evaluate~\algo~on multiple benchmark datasets under various noise types as well as real-world tasks. Experimental results demonstrate the superiority of ~\algo~over the state-of-the-art methods.
\end{enumerate}

The rest of the paper is organized as follows. First, we introduce some related work. Then, we present the studied learning problem and the proposed framework. Furthermore, we experimentally analyze the proposed method. Finally, we conclude this paper.

\section{Related Work}

\textbf{Learning from Noisy Labels} is a heavily studied problem. Many methods attempt to rectify the loss function, which can be categorized into two types. The first type treats samples equally and rectifies the loss by either removing or relabeling noisy samples \cite{han2019selflearning,pleiss2020identifying,yao2020dualt,Nguyen2020self}. For example, AUM~\cite{pleiss2020identifying} designs a margin-based method for detecting noisy samples by observing that clean samples have a larger margin than noisy samples. TopoFilter~\cite{wu2020topofilter} assumes that clean data is clustered together while noisy samples are isolated. Joint-Optim~\cite{Tanaka2018joint} and PENCIL~\cite{yi2019probabilistic} treat labels as learnable variables, which are jointly optimized along with model parameters. Another type of method learns to reweight samples with higher weights for clean data points~\cite{liu2016importance,ren2018learning,Kim2019nlnl}. Instead of using a fixed weight for all samples, M-correction~\cite{arazo2019unsupervised} uses dynamic hard and soft bootstrapping loss to dynamically reweight training samples. Some recent works resort to early-learning regularization~\cite{liu2020early} and data augmentation~\cite{nishi2021augmentation} to handle noisy labels.

The above methods only consider IND label noise in training datasets. Recently, some works~\cite{wang2018iterative,lee2019rog,sachdeva2021evidentialmix,li2021mopro,li2021learning} propose to handle both IND and OOD noise in training datasets. For instance, ILON~\cite{wang2018iterative} discriminates noise samples by density estimating. MoPro~\cite{li2021mopro} and ProtoMix~\cite{li2021learning} identify IND and OOD noise according to predictive confidence. However, these approaches cannot be directly applied for detecting OOD at test time, and the performance of simply combining with existing OOD detection methods is not satisfactory. In this work, we introduce a new framework that simultaneously learns a classifier and an OOD detector from training data with both IND and OOD noise.

\textbf{OOD Detection} aims to identify test data points that are far from the training distribution. According to whether requiring labels during training time, OOD detection methods can be categorized into supervised learning methods~\cite{lee2018detect,liang2018odin,lee2018mahalanobis,yu2019maximumDiscrepancy,hendrycks2019rotplus} and unsupervised learning methods~\cite{geifman2017selective,golan2018rot,sehwag2021ssd}. For example, ODIN~\cite{liang2018odin} separates IND and OOD samples by using temperature scaling and adding perturbations to the input. Lee et al.~\cite{lee2018mahalanobis} obtains the class conditional Gaussian distributions and calculates confidence score based on Mahalanobis distance. Recently, SSD~\cite{sehwag2021ssd} uses self-supervised learning to extract latent feature representations and Mahalanobis distance to compute the membership score between test data points and IND samples. 

Compared with supervised detectors,~\algo~does not assume the availability of clean datasets which are often difficult to obtain in many real-world applications~\cite{weit2018ml,weit2020tnnls}. Instead,~\algo~can detect OOD examples by training on noisy-labeled datasets.

\begin{figure*}[ht]
    \centering
    \includegraphics[width=0.98\linewidth]{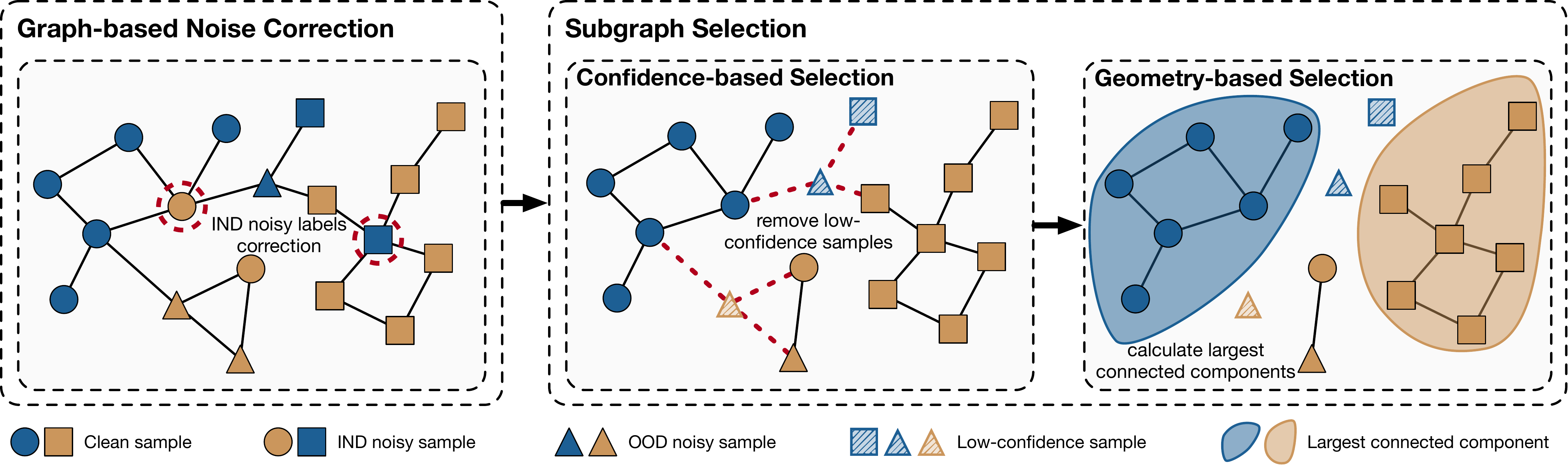}
    \caption{An illustration of graph-based noise correction and subgraph selection in binary classification case.}
    \label{fig:algorithm_illustration}
\end{figure*}

\section{Learning with Open-World Noisy Data}
In this section, we first introduce the studied problem setup and an overview of the proposed noisy graph cleaning framework. Then, we present the proposed framework.

\subsection{Problem Formulation}
Given a training dataset $\mathcal{D}_{train} = \{\x_i, y_i\}_{i=1}^{N}$, where $\x_i$ is an instance feature representation and $y_i \in \mathcal{C} = \{1, \dots, K\}$ is the class label assigned to it. In $\mathcal{D}_{train}$, we assume that the instance-label pair $(\x_i, y_i), 1 \leq i \leq N$ consists of three types. Denote $y_i^{*}$ as the ground-truth label of $\x_i$, a \textbf{ correctly-labeled sample} whose assigned label matches the ground-truth label, i.e., $y_i = y_i^{*}$. An \textbf{ IND mislabeled sample} has an assigned label that does not match the ground-truth label, but the input matches one of the classes in $\mathcal{C}$, i.e., $y_i \neq y_i^{*}$ and $y_i^{*} \in \mathcal{C}$. An \textbf{OOD mislabeled sample} is one where the input does not match the assigned label and other known classes, i.e., $y_i \neq y_i^{*}$ and $y_i^{*} \notin \mathcal{C}$. 
In inference, there are two types of test samples. An \textbf{ IND sample} is one where $\x$ is taken from the distribution of one of the known classes, i.e., $y_i^{*} \in \mathcal{C}$. An \textbf{OOD sample} is the one taken from unknown class distributions, i.e., $y_i^{*} \notin \mathcal{C}$.

\subsection{An Overview of the Proposed Framework}

To address the LOND problem, we present a graph-based framework, named Noisy Graph Cleaning (\algo), which can exploit the relationships among data and learn robust representations from reliable data. Initially, a $k$-NN graph is constructed, where samples are represented as vertices (nodes) in the graph with edges represent similarities between samples. Since labels of samples may be mislabeled, we refer to the resulting graph as \textit{noisy graph}. Then,~\algo~accomplishes noisy graph cleaning in two steps. First, to cope with IND noise,~\algo~refines noisy labels using the proposed soft pseudo-label propagation based on the smoothness assumption~\cite{zhou2003lp,zhu2005semi,iscen2019lpssl}. Second, since OOD samples do not belong to any IND classes, soft pseudo-label propagation is not able to correct their labels. We propose to collect a subset of clean samples to guide the learning of the network. To achieve this goal, a two-stage subgraph selection method is introduced, i.e., confidence-based and geometry-based selection. The confidence-based strategy breaks the edges between nodes with clean labels and noisy labels by removing samples with low-confidence predictions. Then the geometry-based strategy selects nodes that are likely to be clean. Figure~\ref{fig:algorithm_illustration} provides an illustration of the proposed method. We observe that these two selection strategies are indispensable and single application of each one leads to inferior performance. Based on that, we employ devised instance-level and subgraph-level contrastive losses to learn robust representations, which in return can benefit the construction of graph and the subgraph selection. In each training iteration, the graph is re-constructed and noise correction as well as subgraph selection are performed. Then, the selected clean samples are used for the training of DNNs.

\subsection{Graph-based Noise Correction}
The goal of noise correction is to propagate labels on the undirected graph $G = \langle V, E \rangle$ by leveraging similarities between data. $V$ and $E$ denote the set of graph vertices and edges, respectively. In graph $G$, the similarities between vertices are encoded by a weight matrix $\W$. For scalability, we adopt the $k$-NN matrix, which is obtained by:
\begin{equation}\label{equ:knn-weight}
\W_{i j}:=\left\{\begin{array}{ll}
{\left[\z_{i}^{\top} \z_{j}\right]_{+}^{\gamma},} & \text { if } i \neq j \wedge \z_{i} \in \mathbf{N N}_{k}\left(\z_{j}\right) \\
0, & \text { otherwise }
\end{array}\right.
\end{equation}
Here, $\gamma$ is a parameter simply set as $\gamma=1$ in our experiments. $\z_i$ is the latent representation for $\x_i$ and $\mathbf{N N}_{k}$ denotes the $k$ nearest neighbors. To capture high-order graph information, researchers have designed models on the assumption that labels vary smoothly over the edges of the graph~\cite{zhou2003lp,zhu2005semi,iscen2019lpssl,li2021tkde}. In this work, we propose to propagate soft pseudo-labels obtained from the network. 
Denote $\Y = [\y_1, \cdots, \y_N] \in \mathbb{R}^{N\times K}$ as the initial label matrix. We set $\y_i$ to the one-hot label vector of $\x_i$ if $\x_i$ is selected as a clean sample by our method introduced in Section \ref{sec:selection}, otherwise we use model prediction aggregated by temporal ensemble~\cite{laine2016temporal,Nguyen2020self} to initialize it.
Let $\D$ be the diagonal degree matrix for $\W$ with entry $d_{ii}=\sum_j \W_{ij}$, we obtain the refined soft pseudo-labels $\tilde{\Y} = [\tilde{\y}_1, \cdots, \tilde{\y}_N] \in \mathbb{R}^{N\times K}$ by solving the following minimization problem:
\begin{equation}\label{equ:lp}
J(\tilde{\Y}):=\frac{\alpha}{2} \sum_{i, j=1}^{N} \W_{i j}\left\|\frac{\tilde{\y}_{i}}{\sqrt{d_{i i}}}-\frac{\tilde{\y}_{j}}{\sqrt{d_{j j}}}\right\|^{2}+(1-\alpha)\|\Y-\tilde{\Y}\|_{F}^{2}
\end{equation}
In Eq.~(\ref{equ:lp}), all nodes propagate pseudo-labels to their neighbors according to edge weights. $\alpha$ is used to trade-off between information from neighborhoods and vertices themselves and we simply set it to 0.5 in all experiments. This minimization problem can be solved by using conjugate gradient as~\cite{zhu2005semi,iscen2019lpssl}. After obtaining refined soft pseudo-labels, it is common to transform $\tilde{\Y}$ into hard pseudo-labels to guide the training. Specifically, in iteration $t$, the hard pseudo-label for the $i$-th data point is generated by taking the largest prediction score as $\hat{y}_i =  \argmax_k \; \tilde{y}^{(t)}_{ik}$, where $\tilde{y}^{(t)}_{ik}$ represents the $k$-th element in $\tilde{\y}^{(t)}_{i}$.

\subsection{Subgraph Selection }
\label{sec:selection}
When training DNNs with noisy labels, it is observed that clean samples of the same class are usually clustered together in the latent feature space, while noisy samples are pushed away from these clusters~\cite{lee2019rog,wu2020topofilter}. This inspires us to find the connected component with the same class label in the graph for each class. Unfortunately, OOD samples can be similar to some clean samples, leading to undesirable edges in the graph such that nodes corresponding to OOD samples are included in the largest connected component (LCC). To remedy this, we introduce confidence-based selection to remove edges associated with low-confidence nodes because these edges are unreliable. After that, the geometry-based selection is employed to obtain the LCC in subgraphs of each class. 

\textbf{Confidence-based Sample Selection.}
Since low-confidence nodes are more likely to connect OOD nodes to the clusters of clean nodes, we use a sufficiently high threshold $\eta \in [0, 1]$ to select a reliable subset of nodes:
\begin{equation}\label{equ:confidence-selection}
g_i = \left\{\begin{array}{ll}
 1, & \text { if }\tilde{\Y}^{(t)}_{i y_i} >\frac{1}{K} \\
\mathbb{I}\left[\max_k \; \tilde{\Y}^{(t)}_{ik} >\eta \right], & \text { otherwise }
\end{array} \right.
\end{equation}
where $g_i$ is a binary indicator representing the conservation of node $v_i \in V$ when $g_i=1$ and the removal of node $v_i$ when $g_i=0$. Note that we have another condition $\tilde{\Y}^{(t)}_{i y_i} >\frac{1}{K}$ which is complementary to the high-confidence condition inspired by previous works~\cite{li2021mopro,li2021learning}. The reason is that the network may not produce confident predictions in the early phase of training, while it has been observed to first fit the training data with clean labels~\cite{pleiss2020identifying,liu2020early}. Therefore, we incline to treat label $y_i$ as clean if its corresponding prediction score is higher than uniform probability $\frac{1}{K}$, and we set $\hat{y}_i = y_i$. Then we refine graph $G$ based on the indicator $g$ as $\tilde{V} = V \setminus \{v \mid \forall v \in V, g_v = 0\}$ and $\tilde{E} = E \setminus \{e \mid \forall e=\langle e_1, e_2 \rangle \in E, g_{e_1} + g_{e_2} < 2\}$. 
In this way, low-confidence nodes and their corresponding edges are removed from graph $G$ and the resulting graph is denoted by $\tilde{G} = \langle \tilde{V}, \tilde{E} \rangle$. In the modified graph $\tilde{G}$, the connectivity between nodes are more reliable, which facilitates to the geometry-based selection.

\textbf{Geometry-based Sample Selection.}
In graph $\tilde{G}$, we expect that nodes with same labels are connected. Since nodes with noisy labels locate far away from clean ones, more than one connected component may exist for each class. Therefore, we selected the LCC for robustness. Specifically, for the $k$-th class, graph nodes that possess labels of other classes, i.e., $\hat{y}_i \neq k, \forall i \in [N]$, and their adjacent edges $\tilde{G}$ are removed. We denote this as the class-specific subgraph for class $k$ as $\tilde{G}(k)$. Let $\tilde{G}(k)_{lcc}$ be the set of nodes in the LCC of $\tilde{G}(k)$, we obtain a subset of clean samples by $\mathcal{S} = \bigcup_{k=1}^{K} \tilde{G}(k)_{lcc}$. Note that a connected component of $\tilde{G}(k)$ is a subgraph in which any two vertices are connected by edges, and which is not connected to any other vertex in the rest of the graph. In other words, we consider data points belonging to the LCC of the class-specific subgraphs for each class to be clean, since small connected components may contain noisy samples. In practice, we implement disjoint-set data structures to compute the components effectively.

In summary, we identify clean samples by using both predictive confidence and the geometric structure of data:
\begin{equation}\label{equ:final-selection}
g_i = \left\{\begin{array}{ll}
 \mathbb{I} [i \in \mathcal{S} ], & \text { if }\tilde{\Y}^{(t)}_{i y_i} >\frac{1}{K} \\
\mathbb{I}\left[\max_k \; \tilde{\Y}^{(t)}_{ik} >\eta \right] \cdot \mathbb{I} [i \in \mathcal{S} ], & \text { otherwise }
\end{array}\right.
\end{equation}

\subsection{ Subgraph-level Contrastive Learning }
It is noted that exploring the similarities between samples is essentially based on meaningful feature representations. To this end, we take advantage of contrastive learning, which has been successfully used to learn good representations in many tasks~\cite{wu2018unsupervised,chen2020simclr,khosla2020supervised,li2021learning}. The basic idea of contrastive learning is to pull together two embeddings of the same samples, while pushing apart embeddings of other samples. Formally, the instance-level contrastive loss is obtained as follows.
\begin{equation}\label{equ:self-contrastive}
\mathcal{L}^{\mathrm {inst }}=-\sum_{i \in I} \log \frac{\exp \left(\boldsymbol{z}_{i} \cdot \boldsymbol{z}_{j(i)} / \tau_{1}\right)}{\sum_{a \in A(i)} \exp \left(\boldsymbol{z}_{i} \cdot \boldsymbol{z}_{a} / \tau_{1}\right)}
\end{equation}
Here $\boldsymbol{z}_{i}=\operatorname{Proj}\left(\operatorname{Enc}\left(\boldsymbol{x}_{i}\right)\right) \in \mathbb{R}^{D_{P}}$ denotes the $l_2$ normalized feature representation with dimension $D_{P}$, %symbol $\cdot$ denotes the inner (dot) product, 
and $\tau_{1}$ is a scalar temperature parameter. $I$ denotes the set of training samples, $I'$ is another augmented set, and $A(i) = (I \backslash\{i\}) \cup I'$. Different augmentation strategies can be used on $I$ and $I'$ as~\cite{li2021learning}. We use $j(i)$ to denote the index of the other augmented sample of $\x_i$.

However, direct optimization of the instance-level contrastive objective in Eq.~\eqref{equ:self-contrastive} is ineffective, which does not leverage the label information and the geometry of data. To this end, we design a subgraph-level contrastive loss:
\begin{equation}\label{equ:sup-contrastive}
\mathcal{L}^{\mathrm{subgraph}}=\sum_{i \in I} \frac{-1}{|P(i)|} \sum_{p \in P(i)} \log \frac{\exp \left(\boldsymbol{z}_{i} \cdot \boldsymbol{z}_{p} / \tau_{2}\right)}{\sum_{a \in A(i)} \exp \left(\boldsymbol{z}_{i} \cdot \boldsymbol{z}_{a} / \tau_{2}\right)}
\end{equation}
Here $P(i) = \left\{p \in A(i): \hat{y}_{p}=\hat{y}_{i} \wedge g_p + g_i = 2\right\}$, and $|P(i)|$ is its cardinality.
In the calculation of $|P(i)|$, $g_i=1$ indicates that only selected clean samples by~\algo~are used for training. $\tau_{2}$ is another temperature parameter. For each class, samples belonging to the corresponding LCC are pulled together by optimizing Eq.~\eqref{equ:sup-contrastive}. In return, it benefits the clean data selection because more samples of the same class are connected in the $k$-NN graph.

Considering the above definitions and denoting $\mathcal{L}^{\mathrm{ce}}$ as conventional cross-entropy loss, the overall training objective is written as follows.
\begin{equation}
\mathcal{L}=\mathcal{L}^{\mathrm{ce}}+\lambda_{1} \mathcal{L}^{\mathrm{inst}}+\lambda_{2} \mathcal{L}^{\mathrm{subgraph}},
\label{eq:total_loss}
\end{equation}
where hyperparameters $\lambda_{1}$ and $\lambda_{2}$ are simply set to $1$ in all experiments. 
We adopt DNN model as feature extractor $\operatorname{Enc(\cdot)}$ and a linear layer as projector $\operatorname{Proj(\cdot)}$ to generate latent feature representation $\boldsymbol{z}_i$. Another linear layer following the feature extractor is used as classifier. Finally, we train the network by minimizing the total loss in Eq. \eqref{eq:total_loss}.

\subsection{OOD Detection}
By far,~\algo~is able to learn classifiers from data with mixed IND and OOD noise. To fully achieve the goal of LOND, the framework must account for the presence of OOD samples at test time. This motivates us to design a principled way to detect OOD samples by measuring the class-conditional probability. Specifically, given a feature representation learned from~\algo, the class-conditional probability is computed based on the similarity between the latent representation of input $\x$ and the class prototypes $\{\c_k\}_{k=1}^{K}$, where $\c_k$ is the normalized mean embedding for selected clean samples of class $k$, and can be obtained by:
\begin{equation} \c_k = \mathrm{Normalize} ( \frac{1}{\sum_{i\in \mathcal{I}_k}g_i} \sum_{i \in \mathcal{I}_k } g_i \z_i ),
\end{equation}
where $\mathcal{I}_k$ denotes the set of samples for which the corresponding pseudo-labels $\hat{y}_i = k, \forall i \in [N]$. Then, the maximum class-wise similarity is computed as follows.
\begin{equation}
s\left(\x\right):=\max _{k\in [K]} \operatorname{sim}\left(\z, \c_k \right).
\end{equation}
Here $\boldsymbol{z}=\operatorname{Proj}\left(\operatorname{Enc}\left(\boldsymbol{x}\right)\right)$ and $\operatorname{sim}$ stands for any similarity measure. In practice, we measure cosine similarity to compute $s(\x)$. When detecting OOD samples, the lower $s\left (\x\right )$ is, the more likely it is to be an OOD sample. To make hard decisions, the probability threshold $\zeta$ is used. That is, a testing point $\x$ is deemed as OOD if and only if $s(\x) < \zeta$.

\section{Experiments}

In this section, we investigate the performance of the proposed~\algo~on multiple datasets with various label noises. Specifically, we introduce our experiments in three aspects as shown in Table \ref{tab:exp-setup}. We verify the effectiveness of our method in the proposed LOND task and learning with closed-world noisy labels (LCNL) as well as learning from real-world noisy dataset (LRND) tasks in order.  

\setlength{\tabcolsep}{2.7pt}
\begin{table}[htbp]
\centering
\small
\caption{Three types of tasks considered in our experiments.}\label{tab:exp-setup}
\begin{tabular}{c | c  c c }
\toprule
Setup & IND noise in $\mathcal{D}_{train}$ & OOD in $\mathcal{D}_{train}$ & OOD in $\mathcal{D}_{test}$ \\
\midrule
LOND & \checkmark & \checkmark & \checkmark \\
LCNL & \checkmark & \xmark & \xmark\\
LRND & \checkmark & \checkmark & \xmark\\
\bottomrule
\end{tabular}
\end{table}

\textbf{Implementation details.}
For all CIFAR experiments, we train PreAct ResNet-18 network using SGD optimizer with momentum 0.9 and weight decay $5\cdot 10^{-4}$. The initial learning rate is set to 0.15 and cosine decay schedule is used. The batch size is set to 512 and the dimension of projector layer is set to 64. For CIFAR-10 experiments, we use $k=30$ for sym. noise and $k=10$ for asym. noise, warmup with cross-entropy loss for 5 epochs. For CIFAR-100 experiments, we set $k=200$ and warmup for 30 epochs. The network is trained for 300 epochs. Mixup~\cite{zhang2017mixup} and AugMix~\cite{hendrycks2020augmix} are used as data augmentation. We provide detailed experimental settings in the supplementary material.

\setlength{\tabcolsep}{10pt}
\begin{table*}[h]
\centering
\small
\caption{Test accuracy (\%) under mixed IND and OOD noise compared with state-of-the-art LNL methods. 50\% sym. IND noise is injected into dataset. We run methods three times with different seeds and report the mean and the standard deviation.}\label{exp:ind-ood-classification}
\begin{tabular}{ c | c | c | c c c c c}
\toprule
IND dataset & OOD dataset & \# OOD & CE & RoG~\cite{lee2019rog} & ILON~\cite{wang2018iterative} &  DivideMix~\cite{li2020dividemix} & Ours\\
\midrule
\multirow{6}{*}{\shortstack{CIFAR-10}}
 & \multirow{2}{*}{\shortstack{CIFAR-100}} 
 & 10k & 53.36$_{\pm \text{0.92}}$ & 63.01$_{\pm \text{0.46}}$ & 75.17$_{\pm \text{1.50}}$ & 92.73$_{\pm \text{0.27}}$ & \bf 93.69$_{\pm \text{0.09}}$ \\
 && 20k & 50.73$_{\pm \text{0.80}}$ & 62.56$_{\pm \text{1.76}}$ & 74.85$_{\pm \text{1.61}}$ & 92.26$_{\pm \text{0.13}}$ & \bf 92.31$_{\pm \text{0.29}}$  \\
 \cmidrule{2-8}
 & \multirow{2}{*}{\shortstack{TinyImageNet}}
 & 10k & 51.85$_{\pm \text{1.09}}$ & 61.69$_{\pm \text{1.18}}$ & 75.93$_{\pm \text{1.13}}$ & \bf 94.08$_{\pm \text{0.18}}$ & 93.73$_{\pm \text{0.36}}$ \\
 && 20k & 52.32$_{\pm \text{1.41}}$ & 63.15$_{\pm \text{1.13}}$ & 74.63$_{\pm \text{0.74}}$ & \bf 93.83$_{\pm \text{0.08}}$ & 93.54$_{\pm \text{0.21}}$ \\
 \cmidrule{2-8}
 & \multirow{2}{*}{\shortstack{Places-365}}
 & 10k & 54.06$_{\pm \text{0.53}}$ & 64.21$_{\pm \text{0.27}}$ & 76.17$_{\pm \text{0.90}}$ & 93.81$_{\pm \text{0.33}}$ & \bf 94.18$_{\pm \text{0.09}}$ \\
 & & 20k & 55.30$_{\pm \text{1.31}}$ & 63.52$_{\pm \text{1.73}}$ & 76.36$_{\pm \text{1.26}}$ & 93.59$_{\pm \text{0.07}}$ & \bf 93.67$_{\pm \text{0.22}}$ \\
\midrule
\multirow{4}{*}{\shortstack{CIFAR-100}}
 & \multirow{2}{*}{\shortstack{TinyImageNet}}
 & 10k & 37.01$_{\pm \text{0.40}}$ & 52.65$_{\pm \text{0.30}}$ & 51.43$_{\pm \text{0.29}}$ & 70.38$_{\pm \text{0.09}}$ & \bf 74.57$_{\pm \text{0.23}}$  \\
 && 20k & 34.55 $_{\pm \text{0.55}}$ & 50.40 $_{\pm \text{0.44}}$ & 50.14$_{\pm \text{0.66}}$ & 69.89$_{\pm \text{0.25}}$ & \bf 73.49$_{\pm \text{0.11}}$ \\
 \cmidrule{2-8}
 & \multirow{2}{*}{\shortstack{Places-365}}
 & 10k & 37.53$_{\pm \text{0.54}}$ & 52.43$_{\pm \text{0.03}}$ & 50.74$_{\pm \text{0.65}}$ & 70.01$_{\pm \text{0.11}}$ & \bf 74.89$_{\pm \text{0.21}}$ \\
 && 20k & 34.54$_{\pm \text{0.18}}$ & 50.32$_{\pm \text{0.29}}$ & 49.87$_{\pm \text{0.46}}$ & 69.84$_{\pm \text{0.15}}$ & \bf 73.44$_{\pm \text{0.35}}$ \\
\bottomrule
\end{tabular}
\end{table*}

\setlength{\tabcolsep}{6.5pt}
\begin{table*}[h]
\centering
\small
\caption{AUROC (\%) comparison with state-of-the-art OOD detectors. 50\% sym. IND noise is injected into training dataset. 20k and 10k OOD samples are added into training set and test set, respectively. $^+$ indicates supervised detection methods.}\label{exp:ind-ood-detection}
\begin{tabular}{ c | c | c c c c c c c c}
\toprule
IND dataset & OOD dataset & MSP\cite{hendrycks2017msp}$^{+}$ & ODIN\cite{liang2018odin}$^{+}$ & MD\cite{lee2018mahalanobis}$^{+}$ & Rot\cite{golan2018rot} & Rot\cite{hendrycks2019rotplus}$^{+}$ & SSD\cite{sehwag2021ssd} & SSD\cite{sehwag2021ssd}$^{+}$ & Ours\\
\midrule
\multirow{3}{*}{\shortstack{CIFAR-10}}
 & CIFAR-100 & 69.91 & 65.40 & 64.45 & 63.84 & 60.25 & 68.42 & 55.88 &\bf 90.37 \\
 & TinyImageNet & 70.12 & 67.31 & 77.55 & 68.87 & 64.64 & 75.51 & 60.52 &\bf 94.18 \\
 & Places-365 & 71.08 & 71.12 & 70.83 & 50.42 & 69.35 & 77.11 & 62.30 & \bf 94.31 \\
\midrule
\multirow{2}{*}{\shortstack{CIFAR-100}} 
 & TinyImageNet & 86.59 & 91.36 & 67.33 & 58.63 & 57.40 & 68.50 & 65.48 & \bf 94.24  \\
 & Places-365 & 85.82 & 89.93 & 68.08 & 44.85 & 59.90 & 68.97 & 76.16 & \bf 91.20 \\
\bottomrule
\end{tabular}
\end{table*}

\begin{table}[h]
\setlength{\tabcolsep}{2.7pt}
\small
\centering
\caption{F-measure comparison with DivideMix (DM) combined with OOD detection methods. 50\% sym. IND noise is injected into training set, 20k and 10k OOD samples are added into training set and test set, respectively.}
\begin{tabular}{ c | c | c c c c c}
\toprule
IND dataset & OOD dataset & DM & MSP & ODIN & MD & Ours\\
\midrule
\multirow{3}{*}{\shortstack{CIFAR-10}}
 & CIFAR-100 & 0.632 & 0.698 & 0.681 & 0.635 & \bf 0.838  \\
 & TinyImageNet & 0.638 & 0.726 & 0.707 & 0.702 & \bf 0.875 \\
 & Places-365 & 0.637 & 0.717 & 0.705 & 0.651 & \bf 0.887 \\
\midrule
\multirow{2}{*}{\shortstack{CIFAR-100}}
 & TinyImageNet & 0.516 & 0.687 & 0.705 & 0.526 & \bf 0.773  \\
 & Places-365 & 0.519 & 0.685 & 0.696 & 0.541 & \bf 0.731 \\
\bottomrule
\end{tabular}
\label{exp:ind-ood-detection-fmeasure}
\end{table}

\setlength{\tabcolsep}{3.5pt}
\begin{table*}[ht]
\small
\centering
\caption{Test accuracy (\%) under controlled IND label noise compared with state-of-the-art methods on CIFAR-10 and CIFAR-100 datasets. We run our method three times with different random seeds and report the mean and the standard deviation. Results for baseline methods are copied from \cite{li2020dividemix,li2021learning}} 
\begin{tabular}{l|c c c c | c |c c c c}
\toprule
Dataset & \multicolumn{5}{ c |}{CIFAR-10} & \multicolumn{4}{ c }{CIFAR-100} \\
\midrule
Noise type & \multicolumn{4}{ c |}{Sym.} & Asym. & \multicolumn{4}{ c }{Sym.} \\
\midrule
Noise level & 20\% & 50\% & 80\% & 90\% & 40\% & 20\% & 50\% & 80\% & 90\% \\
\midrule
Cross-Entropy & 82.7 & 57.9 & 26.1 & 16.8 & 85.0 & 61.8 & 37.3 & 8.8 & 3.5 \\
F-correction~\cite{patrini2017making} & 83.1 & 59.4 & 26.2 & 18.8 & 87.2 & 61.4 & 37.3 & 9.0 & 3.4 \\
Co-teaching+~\cite{yu2019does} & 88.2 & 84.1 & 45.5 & 30.1 & - & 64.1 & 45.3 & 15.5 & 8.8 \\
Mixup~\cite{zhang2017mixup} & 92.3 & 77.6 & 46.7 & 43.9 & - & 66.0 & 46.6 & 17.6 & 8.1 \\
P-correction~\cite{yi2019probabilistic} & 92.0 & 88.7 & 76.5 & 58.2 & 88.5 & 68.1 & 56.4 & 20.7 & 8.8 \\
Meta-Learning~\cite{li2019learningtolearn} & 92.0 & 88.8 & 76.1 & 58.3 & 89.2 & 67.7 & 58.0 & 40.1 & 14.3 \\
M-correction~\cite{arazo2019unsupervised} & 93.8 & 91.9 & 86.6 & 68.7 & 87.4 & 73.4 & 65.4 & 47.6 & 20.5 \\
DivideMix~\cite{li2020dividemix} & 95.0 & 93.7 & \bf 92.4 & 74.2 & 91.4 & 74.8 & 72.1 & 57.6 & 29.2\\
ProtoMix~\cite{li2021learning} & 95.8 & 94.3 & \bf 92.4 & 75.0 & \bf 91.9 & 79.1 & 74.8 & 57.7 & 29.3 \\
\midrule
Ours & \bf 95.88\scriptsize $\pm \text{0.13}$ & \bf 94.54\scriptsize $\pm \text{0.35}$ & 91.59\scriptsize $\pm \text{0.31}$ & \bf 80.46\scriptsize $\pm \text{1.97}$ & 90.55\scriptsize $\pm \text{0.29}$ & \bf 79.31\scriptsize $\pm \text{0.35}$ & \bf 75.91\scriptsize $\pm \text{0.39}$ & \bf 62.70\scriptsize $\pm \text{0.37}$ & \bf 29.76\scriptsize $\pm \text{0.85}$ \\
\bottomrule
\end{tabular}
\label{exp:ind-noise}
\end{table*}

\subsection{ Learning with Open-World Noisy Data }

To investigate the effectiveness of~\algo, we test it under mixed IND and OOD label noise. In this setup, we report both classification and OOD detection performance to show that~\algo~can learn a good classifier and OOD detector simultaneously. We use CIFAR-10 and CIFAR-100 as IND datasets, and TinyImageNet and Places-365 as the OOD datasets. We first add 50\% symmetric IND noise. Then, additional samples are randomly selected from the OOD datasets to form the training dataset. It is noted that the CIFAR-100 dataset is also used as one of the OOD datasets when CIFAR-10 is treated as the IND dataset.

First, we present the classification performance in Table~\ref{exp:ind-ood-classification}. We compare~\algo~with the cross-entropy baseline and three recent methods for LNL, i.e., ILON~\cite{wang2018iterative}, RoG~\cite{lee2019rog} and DivideMix~\cite{li2020dividemix}.

ILON reweights samples based on the outlier measurement. RoG uses an ensemble of generative classifiers built from features extracted from multiple layers of the pretrained model. DivideMix is the state-of-the-art method for LNL. We report the results of DivideMix without ensemble for a fair comparison. The number of OOD samples in training datasets is set to either 10k or 20k. We can see that~\algo~and DivideMix significantly outperform the other three methods. On CIFAR-10,~\algo~achieves better or on par performance compared with DivideMix. On CIFAR-100,~\algo~obtains an average performance gain of $\sim$4\%. This demonstrates the superiority of~\algo~in classification.

Next, we present the OOD detection performance using AUROC in Table~\ref{exp:ind-ood-detection} following~\cite{hendrycks2017msp} and open-set classification performance~\cite{bendale16cvpr} using F-measure in Table~\ref{exp:ind-ood-detection-fmeasure} as the metric. Since different OOD detectors need particularly tuned probability thresholds $\zeta$, for fair comparison, we search the best $\zeta$ for all methods. Noted that LOND has not been studied before, we hence combine one of the best LNL methods DivideMix with leading OOD detectors including MSP~\cite{hendrycks2017msp}, ODIN~\cite{liang2018odin} and Mahalanobis distance (MD)~\cite{lee2018mahalanobis} for comparisons. We also compare with recent OOD detection methods, Rot~\cite{golan2018rot,hendrycks2019rotplus} and SSD~\cite{sehwag2021ssd}, which cannot be simply combined with DivideMix and need separate training. From the results, it can be seen that most comparison methods perform significantly worse than~\algo. In terms of AUROC,~\algo~obtains performance gains over 17.2\% on CIFAR-10 and 1.27\% on CIFAR-100. Regarding F-measure,~\algo~outperforms other methods by at least 14\% on CIFAR-10 and 3.5\% on CIFAR-100. In supplementary material, we conduct comprehensive comparisons with another recent method for LNL, i.e., ProtoMix~\cite{li2021learning}, due to limited space. We also provide further analysis to show that our method is robust to the selection of $\zeta$. 

\subsection{Learning with Closed-World Noisy Labels}

In addition to the LOND task, we test~\algo~in the conventional closed-world noisy label setup. We conduct experiments under controlled IND noise using the CIFAR-10 and CIFAR-100 datasets. To validate the efficacy of~\algo, we compare it with many existing methods, including Meta-Learning~\cite{li2019learningtolearn}, P-correction~\cite{yi2019probabilistic}, M-correction~\cite{arazo2019unsupervised}, DivideMix~\cite{li2020dividemix}, and ProtoMix~\cite{li2021learning}. Following commonly used LNL setups~\cite{arazo2019unsupervised,li2020dividemix}, we run algorithms under asymmetric noise and symmetric noise with different noise levels. The noise level for symmetric noise ranges from 20\% to 90\% where it consists of randomly selecting labels for a percentage of the training data using all possible labels (i.e., the true label could be randomly retained). The noise level for asymmetric noise is set to 40\%.

As Table~\ref{exp:ind-noise} shown, in most cases, our method outperforms recent methods particularly designed for closed-world noisy label problems. This indicates the superiority and robustness of~\algo.

\subsection{Learning from Real-World Noisy Dataset}
We test the performance of our method on real-world dataset WebVision~\cite{webvision} which contains noisy-labeled images collected from Flickr and Google. Similar to previous work~\cite{li2020dividemix}, we perform experiments on the first 50 classes.

\setlength{\tabcolsep}{9pt}
\begin{table}[htbp]
\centering
\small
\caption{Accuracy (\%) on WebVision-50 and ILSVRC2012 validation sets. Results of baselines are from \cite{chen2019understanding,li2020dividemix,liu2020early}.}
\begin{tabular}{l|c|c|c|c}
\toprule
\multirow{2}{*}{Method}
& \multicolumn{2}{ c |}{WebVision} & \multicolumn{2}{ c }{ILSVRC12} \\
\cmidrule{2-5}
 & top-1 & top-5 & top-1 & top-5 \\
\midrule
F-correction~\cite{patrini2017making} & 61.12 & 82.68 & 57.36 & 82.36 \\
Decoupling~\cite{malach2017decoupling} & 62.54 & 84.74 & 58.26 & 82.26 \\
D2L~\cite{ma2018dimensionality} & 62.68 & 84.00 & 57.80 & 81.36  \\
MentorNet~\cite{jiang2018mentornet} & 63.00 & 81.40 & 57.80 & 79.92 \\
Co-teaching~\cite{han2018co} & 63.58 & 85.20 & 61.48 & 84.70 \\
Iterative-CV~\cite{chen2019understanding} & 65.24 & 85.34 & 61.60 & 84.98 \\
DivideMix~\cite{li2020dividemix} & 77.32 & 91.64 & \bf 75.20 & 90.84 \\
ELR+~\cite{liu2020early} & 77.78 & 91.68 & 70.29 & 89.76 \\
\midrule
Ours & \bf 79.16 & \bf 91.84 & 74.44 & \bf 91.04   \\
\bottomrule
\end{tabular}
\label{exp:webvision}
\end{table}

We report comparison results in Table~\ref{exp:webvision}, measuring top-1 and top-5 accuracy on WebVision validation set and ImageNet ILSVRC12 validation set. ~\algo~consistently outperforms competing methods in most cases, which verifies the efficacy of our method on real-world noisy label task.

\setlength{\tabcolsep}{4pt}
\begin{table}[h]
\centering
\small

\caption{Ablation study. GNC denotes graph-based noise correction. CS denotes confidence-based selection and GS denotes graph-based selection. For experiments whose noise type is OOD, Places-365 is used as OOD dataset and 50\% sym. IND noise is injected into training set.}

\begin{tabular}{l|c | c | c | c | c  c }
\toprule
Dataset & \multicolumn{3}{ c |}{CIFAR-10} & \multicolumn{3}{ c }{CIFAR-100} \\
\midrule
Noise type & OOD & Sym. & Asym. & OOD & \multicolumn{2}{ c  }{Sym.} \\
\midrule
Noise level & 20k & 50\%  & 40\% & 20k & 50\% & 80\% \\
\midrule
w/o GNC & 92.13 & 94.32 & 85.85 & 72.85 & 74.20 & 55.56  \\
w/o CS & 87.20 & 92.44 & 89.68 & 63.78 & 73.22 & 37.82 \\
w/o GS & 86.55 & 85.59 & 81.17 & 65.34 & 67.18 & 35.16  \\
w/o $\mathcal{L}^{\mathrm{inst}}$ & 92.45 & 94.02 & 82.67 & 71.38 & 73.30 & 51.59   \\
w/o $\mathcal{L}^{\mathrm{subgraph}}$ & 70.39 & 85.12 & 79.17 & 55.12 & 58.06 & 41.42  \\
w/o mixup & 89.51 & 90.73 & 84.24 & 66.93 & 68.06 & 42.59 \\
w/o AugMix & 93.62 & 94.53 & 89.39 & 71.49 & 75.18 & 61.75 \\
\midrule
Ours & \bf 93.67 & \bf 94.54 & \bf 90.55 & \bf 73.44 & \bf 75.91 & \bf 62.70 \\
\bottomrule
\end{tabular}
\label{exp:ablation}
\end{table}

\begin{figure*}[h]
    \centering
    \begin{subfigure}[b]{0.24\textwidth}
        \centering
        \includegraphics[width=\linewidth]{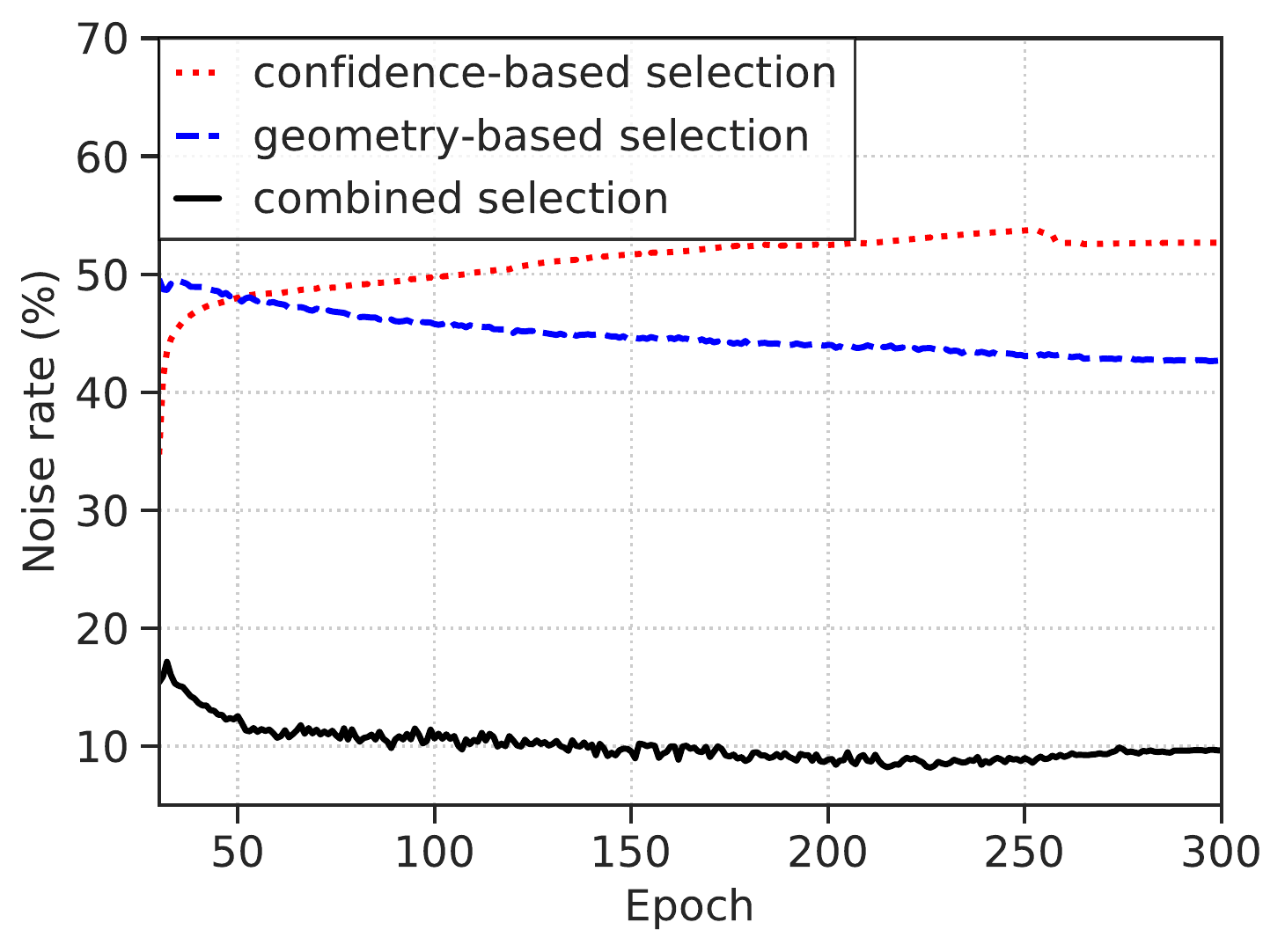}
        \caption{Noise rate in selected samples} \label{fig:noise-rate}
    \end{subfigure}
    \begin{subfigure}[b]{0.24\textwidth}
        \centering
        \includegraphics[width=\linewidth]{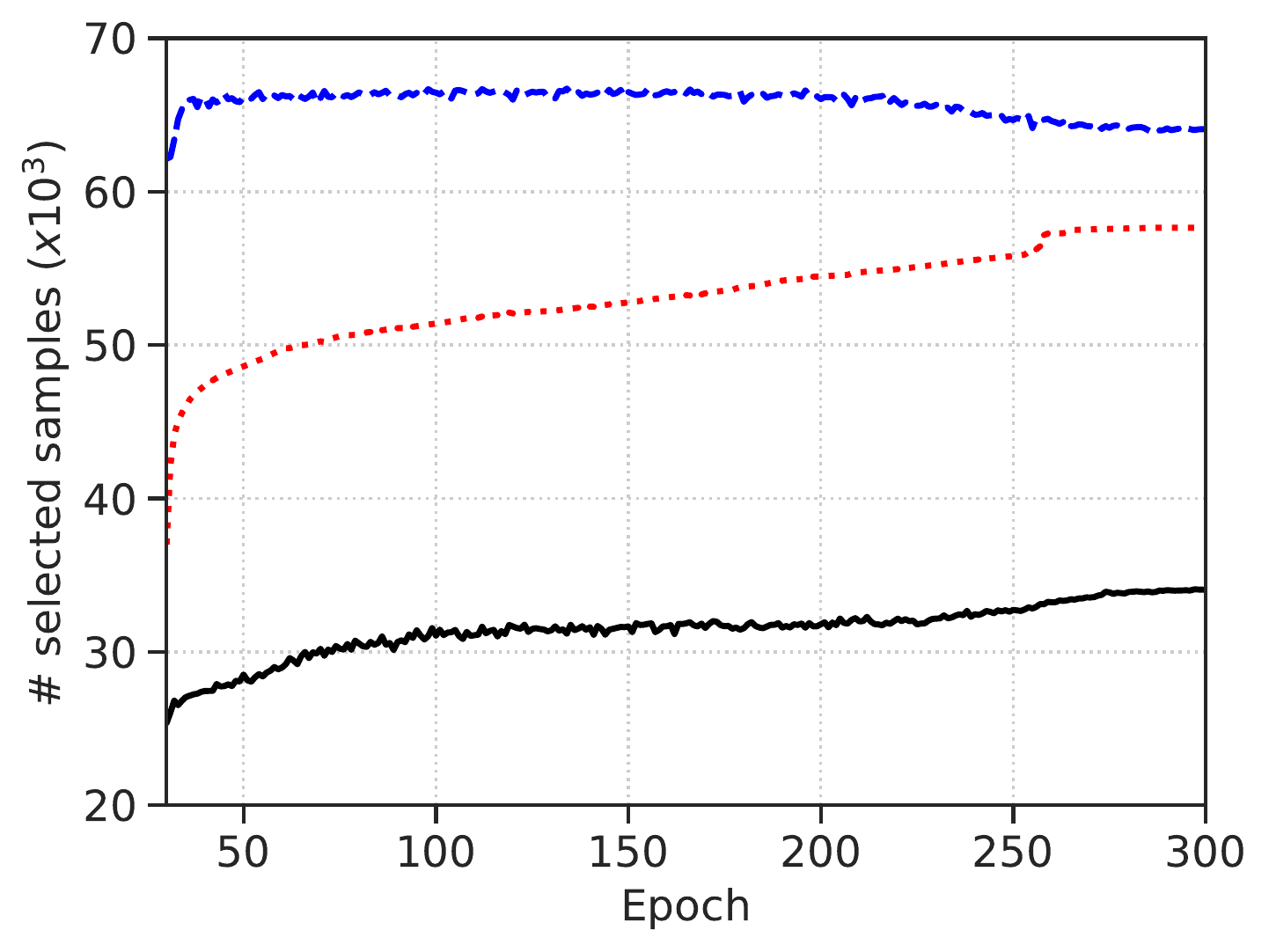}
        \caption{Selected sample size}
        \label{fig:sample-size}
    \end{subfigure}
    \begin{subfigure}[b]{0.24\textwidth}
        \centering
        \includegraphics[width=\linewidth]{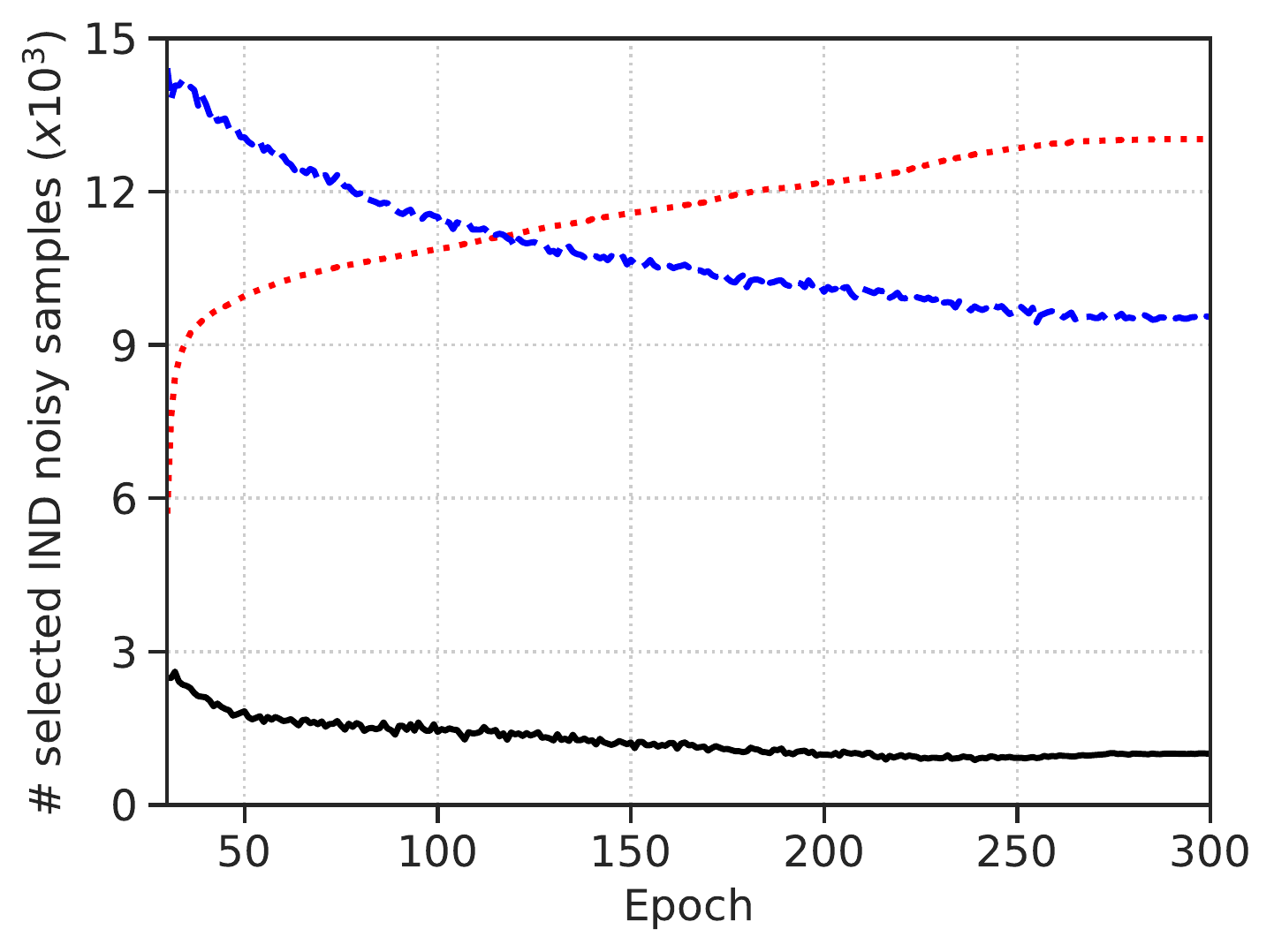} 
        \caption{Selected IND noise}\label{fig:ind-size}
    \end{subfigure}
     \begin{subfigure}[b]{0.24\textwidth}
        \centering
        \includegraphics[width=\linewidth]{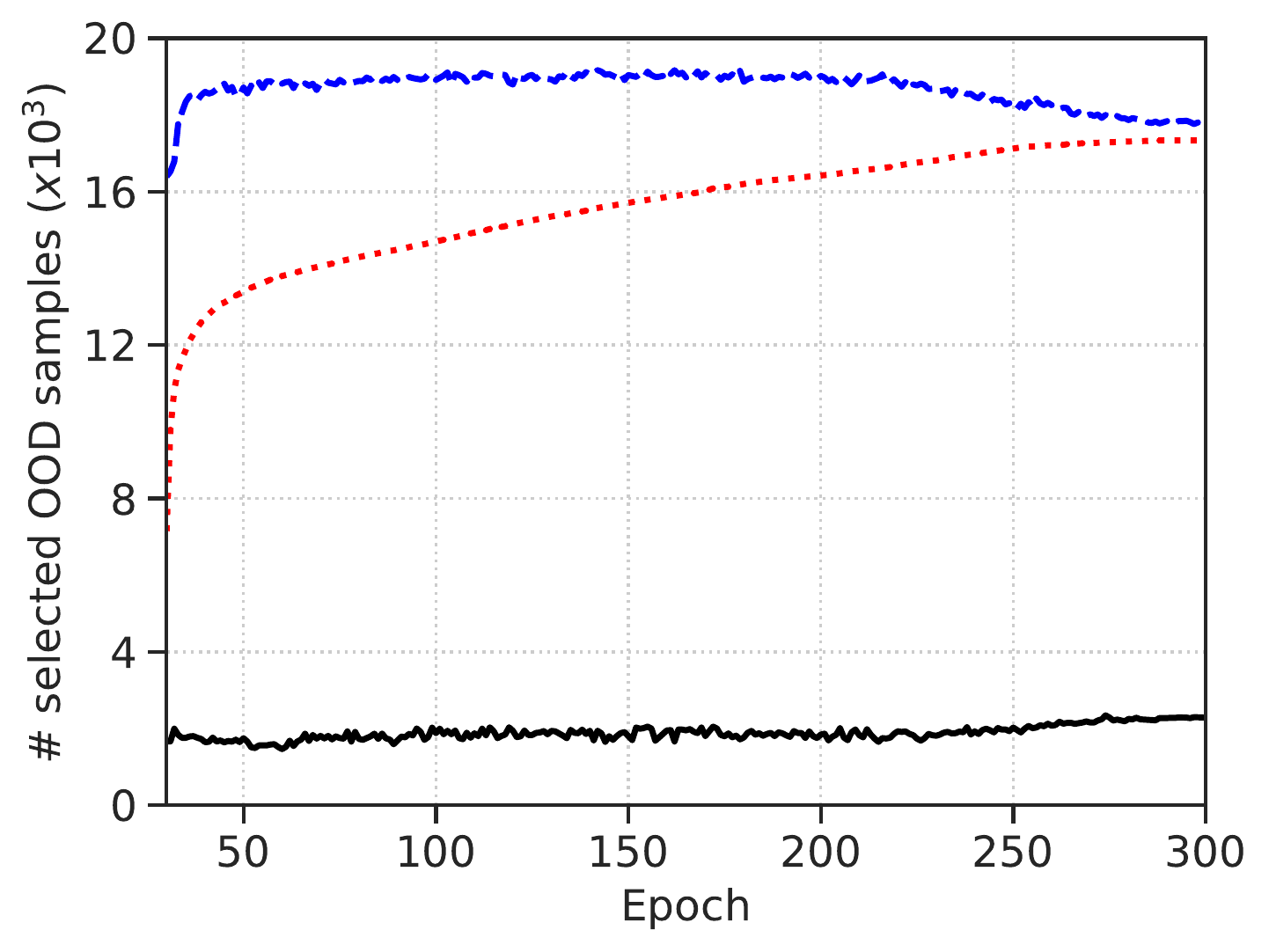}
    \caption{Selected OOD noise}\label{fig:ood-size}
    \end{subfigure}
    \caption{Analysis of subgraph selection under 50\% IND noise (CIFAR-100) and 20k OOD noise (Places-365).}
    \label{fig:sample-selection-performance}
\end{figure*}

\subsection{Ablation Studies and Discussion}
To better understand~\algo, we examine the impact of each component of~\algo~in Table~\ref{exp:ablation}. It can be observed that all components contribute to the efficacy of~\algo. In particular, the two strategies in subgraph selection and the subgraph-level contrastive learning serve as the most important parts in our framework, without which the performance deteriorates severely. The observations validate that confidence-based (CS) and geometry-based selection (GS) can exploit neighborhood information from graph structure effectively. As a result, the test accuracy and OOD detection performance also improve as shown in Figure~\ref{fig:convergence}, demonstrating the good generalization ability of our method. In supplementary material, we also demonstrate the robustness of our method to hyperparameters, i.e., $\eta$ in Eq.~\eqref{equ:confidence-selection} and $k$ which is used to construct the $k$-NN graph.

\textbf{Discussion on subgraph selection.} To further examine the effect of the two subgraph selection strategies, we investigate the impact of each one for selecting clean samples. In Figure~\ref{fig:noise-rate} and Figure~\ref{fig:sample-size}, we can see that the noise rate in the selected data by performing each strategy alone is significantly larger than the combined strategy. 
Figure~\ref{fig:ind-size} and Figure~\ref{fig:ood-size} further show that both IND and OOD noise can be drastically removed by the combined strategy, while merely using one of them has little effect. This is because the confidence-based selection can degrade the connectivity between clean samples and noisy samples such that samples in the largest connected component are clean. Moreover, we divide the nodes into three parts: clean data, IND noise and OOD noise, and analyze the average degrees of nodes in each part after performing confidence-based selection. As demonstrated in Figure~\ref{fig:node_degree}, we find that as the training process progresses, the
average node degree of OOD noisy samples is decreasing, while the average degree of clean samples is increasing. This further validates that confidence-based strategy facilitates the selection of the largest connected component in geometry-based strategy.

\begin{figure}[h]
    \centering
    \begin{subfigure}[b]{0.235\textwidth}
        \centering
        \includegraphics[width=1\linewidth]{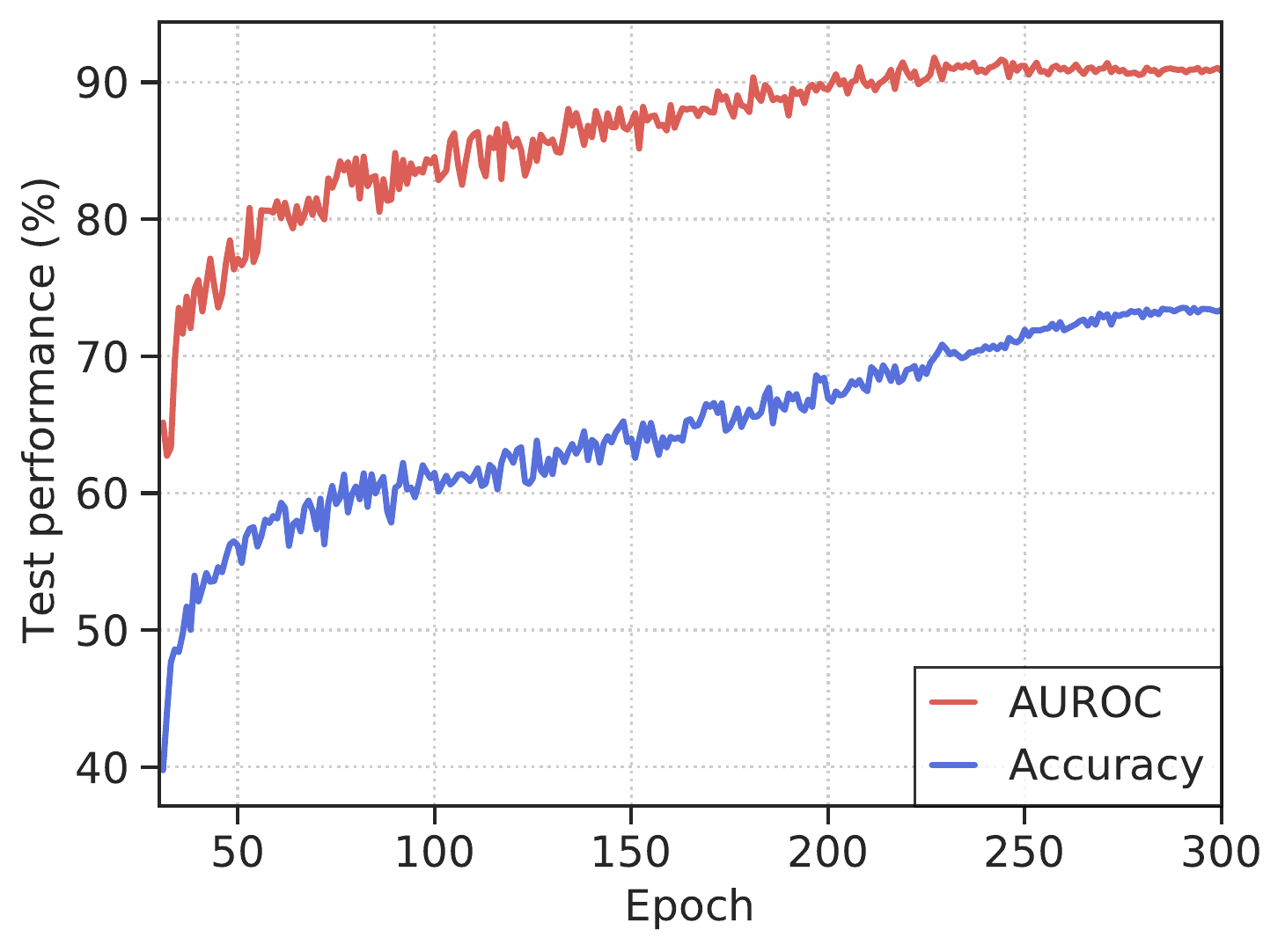}
        \caption{Convergence}\label{fig:convergence}
    \end{subfigure}
    \begin{subfigure}[b]{0.235\textwidth}
        \centering
         \includegraphics[width=1\linewidth]{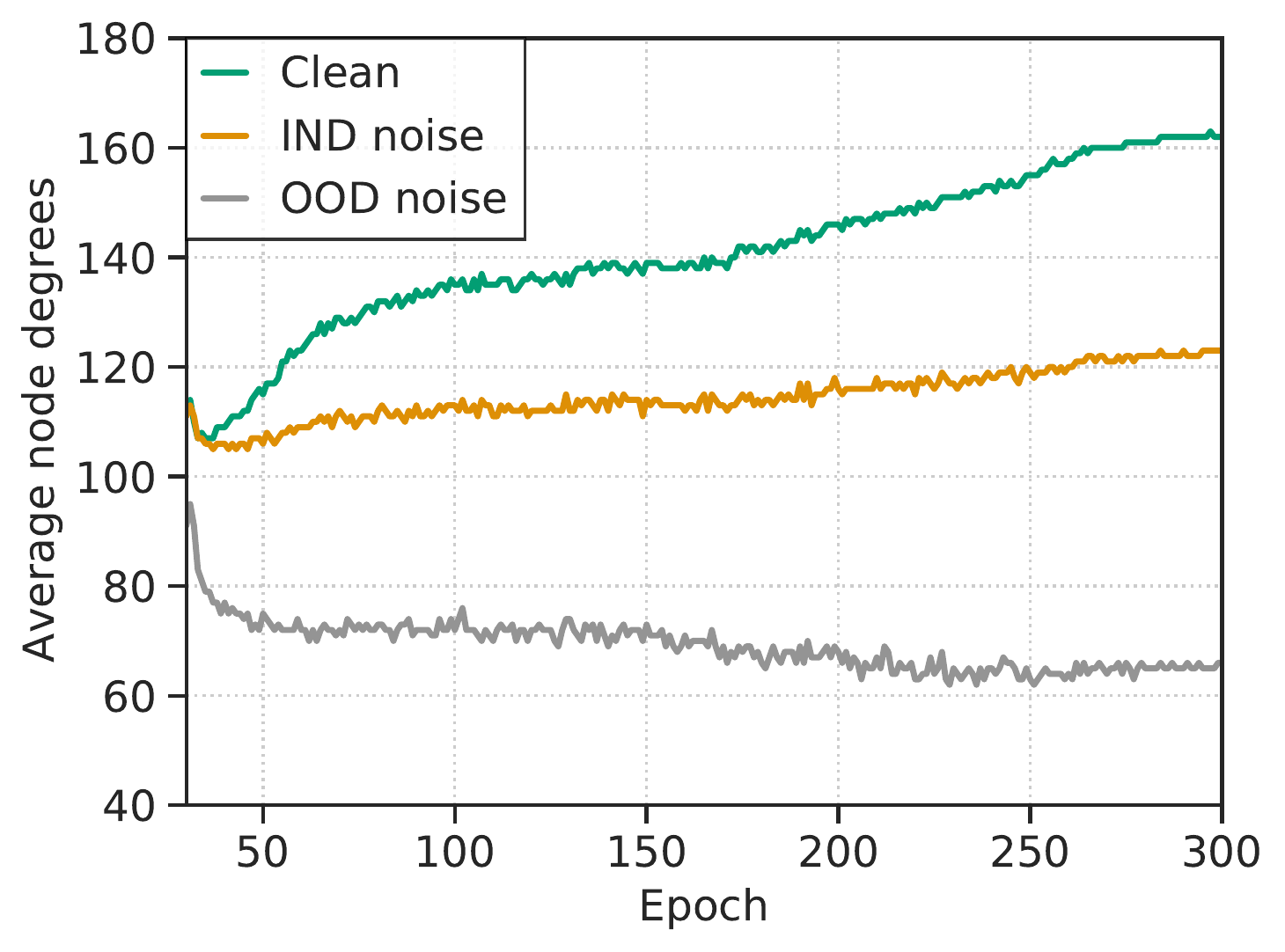}
        \caption{Node degree}\label{fig:node_degree}
    \end{subfigure}  
    \caption{Visualization for convergence of our method and node degrees under 50\% IND noise (CIFAR-100) and 20k OOD (Places-365).}
\end{figure}

\section {Conclusion}
In this paper, we study a realistic problem where the training dataset contains both IND and OOD noise, and the presence of OOD samples at test time. To address this problem, we introduce a noisy graph cleaning framework that simultaneously performs noise correction and clean data selection based on prediction confidence and geometric structure of data in latent feature space. ~\algo~outperforms many existing methods on different datasets with varying degrees of noise. Our work may motivate researchers in two directions: learning from IND and OOD noisy data is worth further exploration due to its broad range of applications and OOD detection from weakly-labeled datasets is promising.

{
\small
\normalem
\bibliographystyle{ieee_fullname}
\bibliography{egbib}
}

\newpage
%\appendix
\appendixtitleon
\appendixtitletocon

\begin{appendices}

\section{Experimental Details}
In this section, we introduce the experiment details. We first introduce the out-of-distribution (OOD) datasets used in our experiments. Then, we present the experimental settings of our method. Finally, we provide details about the evaluation metrics used for evaluating the classification and OOD detection performance of our method.

\subsection{Out-of-Distribution Datasets}
We use the OOD datasets below in our experiments:
\begin{itemize}
    \item \textbf{TinyImageNet}. The Tiny ImageNet dataset contains 50,000 training images from 200 different classes, which are drawn from the original 1,000 classes of ImageNet. We randomly choose samples from training set and resize each image to 32 × 32.
    \item \textbf{Places-365}. The Places-365 dataset has 365 scene categories and there are 900 images per category in the test set. The OOD samples are randomly chosen from test set of Places-365 and resize to 32 × 32.
\end{itemize}

\subsection{Experimental Setup}
For all CIFAR experiments, we train PreAct ResNet-18 network for 300 epochs using SGD with the momentum 0.9 and weight decay $5\cdot 10^{-4}$. The initial learning rate is set to 0.15 and cosine decay schedule is used. The batch size is set to 512. The dimension of projector layer is set to 64. The temperature parameter is fixed as $\tau_{1}=0.3$ and $\tau_{2}=1.0$. For CIFAR-10 experiments, we use $k=30$ for sym. noise and $k=10$ for asym. noise, warmup with cross-entropy loss without other components for 5 epoch. For all CIFAR-100 experiments, we use $k=200$, warmup for 30 epoch for CIFAR-100 datasets. For parameter $\eta$, in LOND task, we use 0.8 for all experiments, and in closed-world noisy label task, we set it to 0.7 for CIFAR-10 and 0.6 for CIFAR-100.

For Webvision-50 dataset, most of hyperparameters are the same with CIFAR experiments except we set $k=100$, $\eta = 0.8$. We train the inception-resnet v2 model using SGD following prior works. The initial learning rate is set to 0.2 and the batch size is 256. We train the network for 80 epochs and the warmup stage lasts 15 epochs.

\subsection{ Evaluation Metrics }
We use the following three performance metrics to evaluate the performance.
\begin{itemize}
    \item \textbf{Classification Accuracy.} The top-1 classification accuracy is calculated as the mean accuracy over all known (IND) classes. Predictions of data are obtained as the classes with the highest softmax probabilities.
    \item \textbf{AUROC.} AUROC is the Area Under the Receiver Operating Characteristic curve and can be calculated by the area under the TPR against FPR curve.
    \item \textbf{F-measure.} The F-measure (F) is calculated as 2 times the product of precision (p) and recall (r) divided by the sum of p and r:
        \begin{equation}
            F=2 \cdot \frac{p \cdot r}{p+r}.
        \end{equation}
    $p$ is calculated as true positive over the sum of $T_p$ and false positive: 
        \begin{equation}
            p=\frac{T_{p}}{T_{p}+F_{p}}.
        \end{equation}
    $r$ is calculated as $T_p$ over the sum of $T_p$ and false negative:
        \begin{equation}
        r=\frac{T_{p}}{T_{p}+F_{n}}.
        \end{equation}
\end{itemize}

\begin{figure*}
    \centering
    \begin{subfigure}[b]{0.33\textwidth}
        \centering
        \includegraphics[width=1\linewidth]{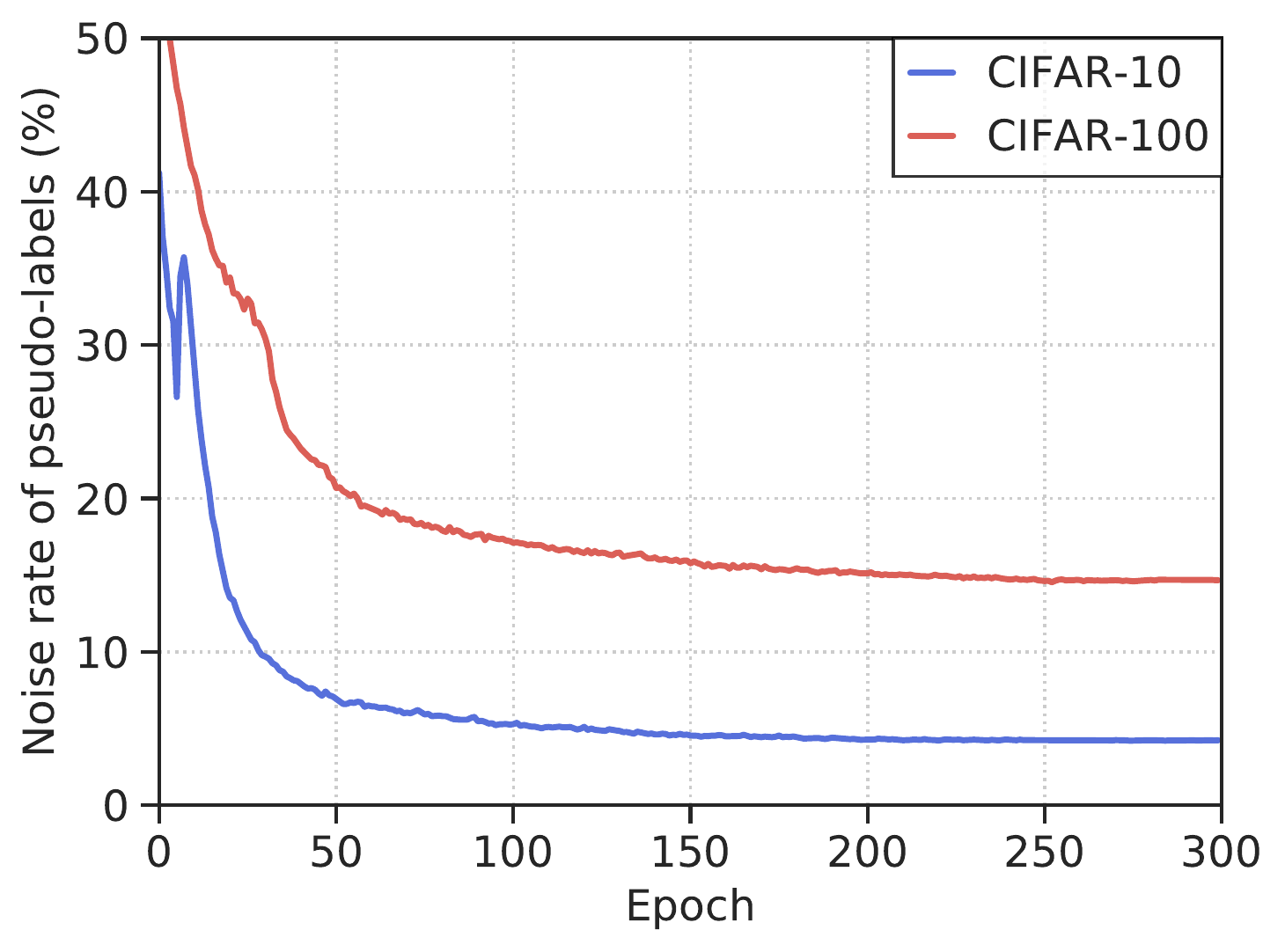}
        \caption{Effectiveness of noise correction.}\label{fig:lp}
    \end{subfigure}
    \begin{subfigure}[b]{0.33\textwidth}
        \centering
        \includegraphics[width=1\linewidth]{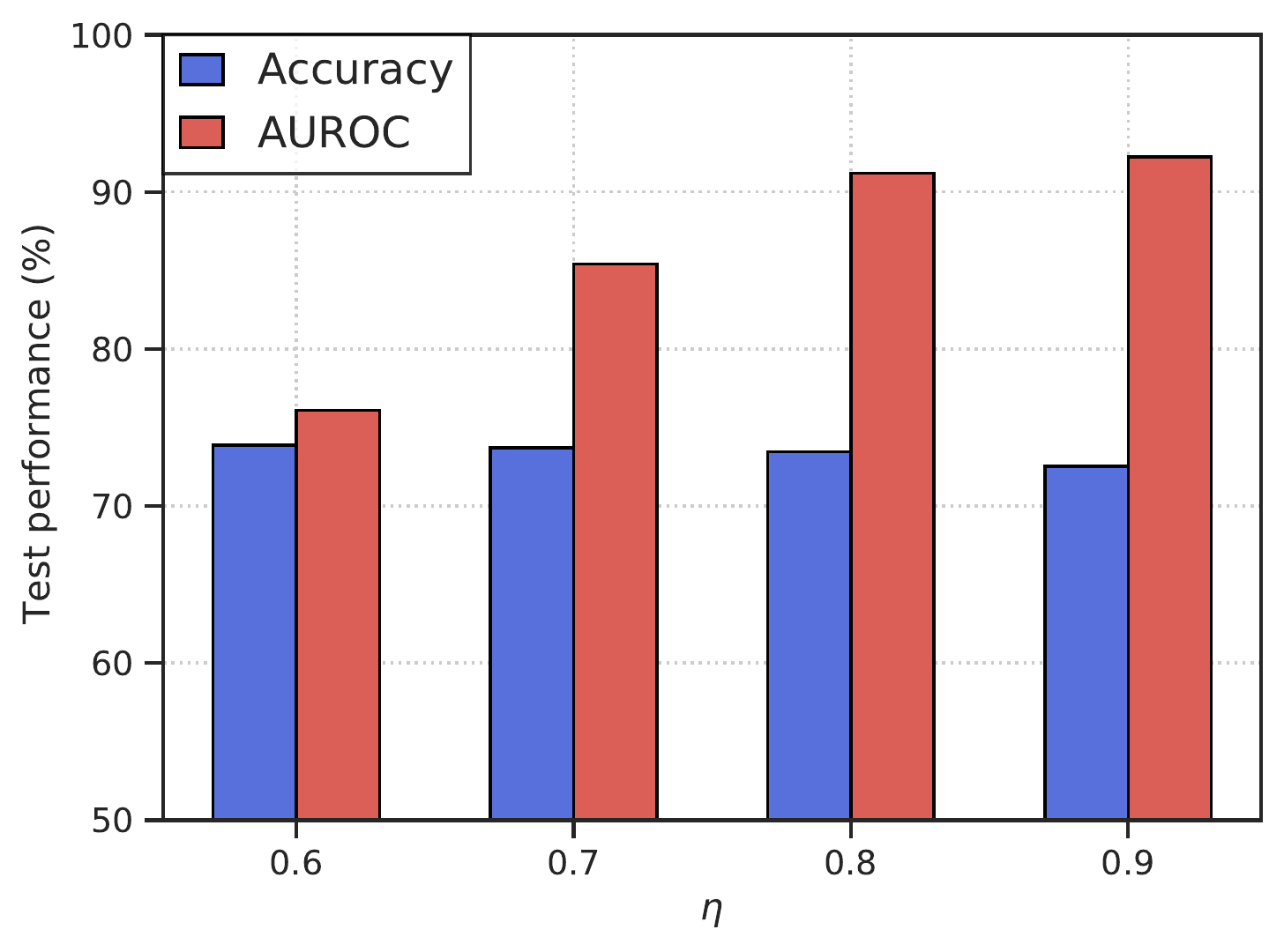}
        \caption{Impact of hyperparameter $\eta$}\label{fig:eta}
    \end{subfigure}
    \begin{subfigure}[b]{0.33\textwidth}
        \centering
         \includegraphics[width=1\linewidth]{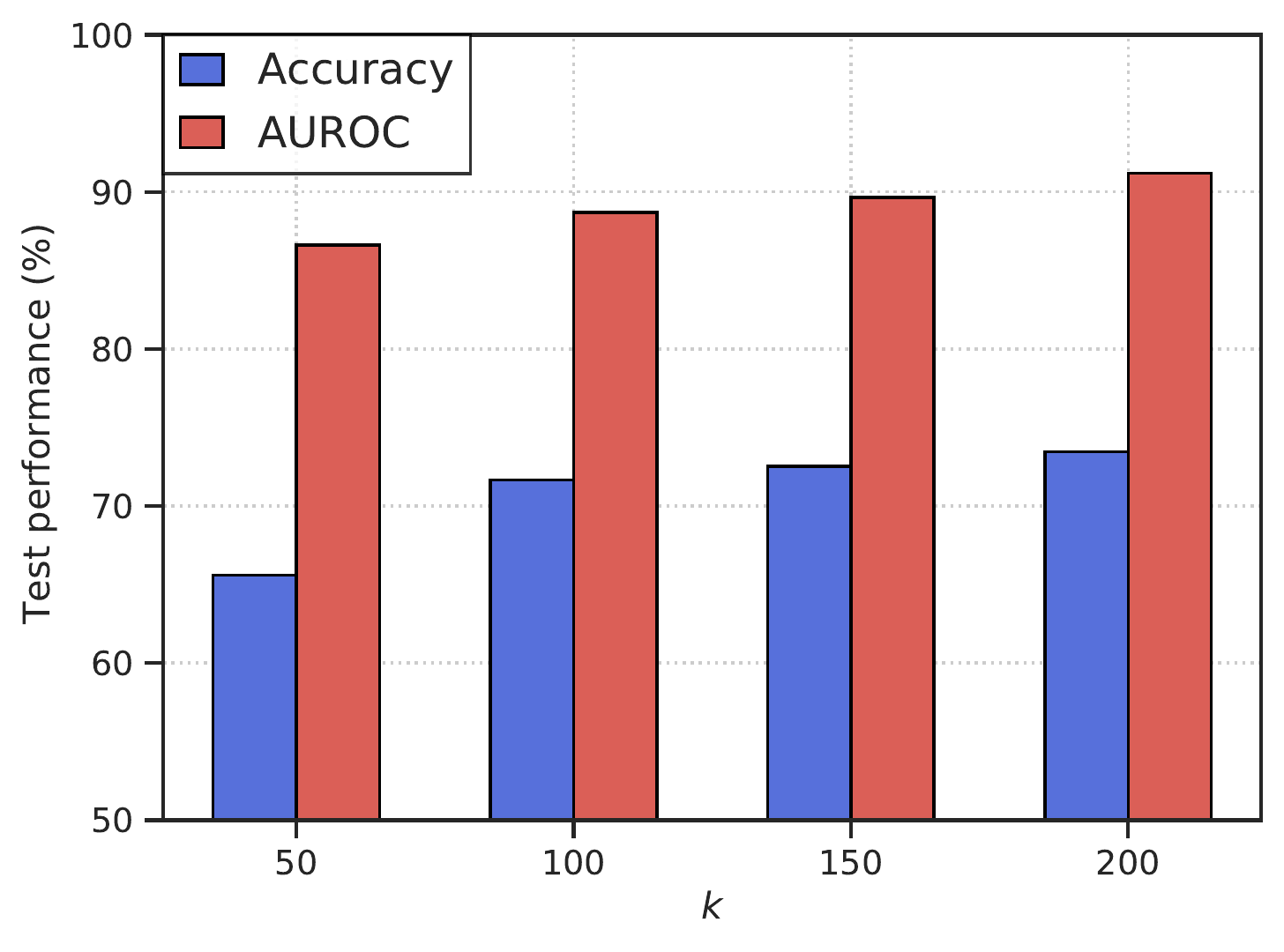}
        \caption{Impact of hyperparameter $k$}\label{fig:k}
    \end{subfigure}  
    \caption{Experimental results. (a)~Effectiveness of noise correction. Both CIFAR-10 and CIFAR-100 datasets are under 50\% sym. noise. (b-c)~Analysis of the impact of hyperparameters under 50\% IND noise (CIFAR-100), 20k and 10k OOD noise (Places-365) in training set and test set, respectively. $\eta$ is for confidence-based selection and $k$ is for $k$-NN graph.}
\end{figure*}

\section{Additional Experimental Results}
In this section, we first show the visualization results of feature representation and subgraph selection, which demonstrate the validity of our methods. Then we present the effectiveness of graph-based noise correction. We also analyze the sensitivity of hyperparameters. In addition, the performance of model ensemble and the impact of AugMix on WebVision-50 is provided. Finally, we compare NGC with recent related work, ProtoMix~\cite{li2021learning} on LOND task.

\begin{figure*}[!h]
    \centering
    \begin{subfigure}[b]{0.33\textwidth}
        \centering
        \includegraphics[width=\linewidth]{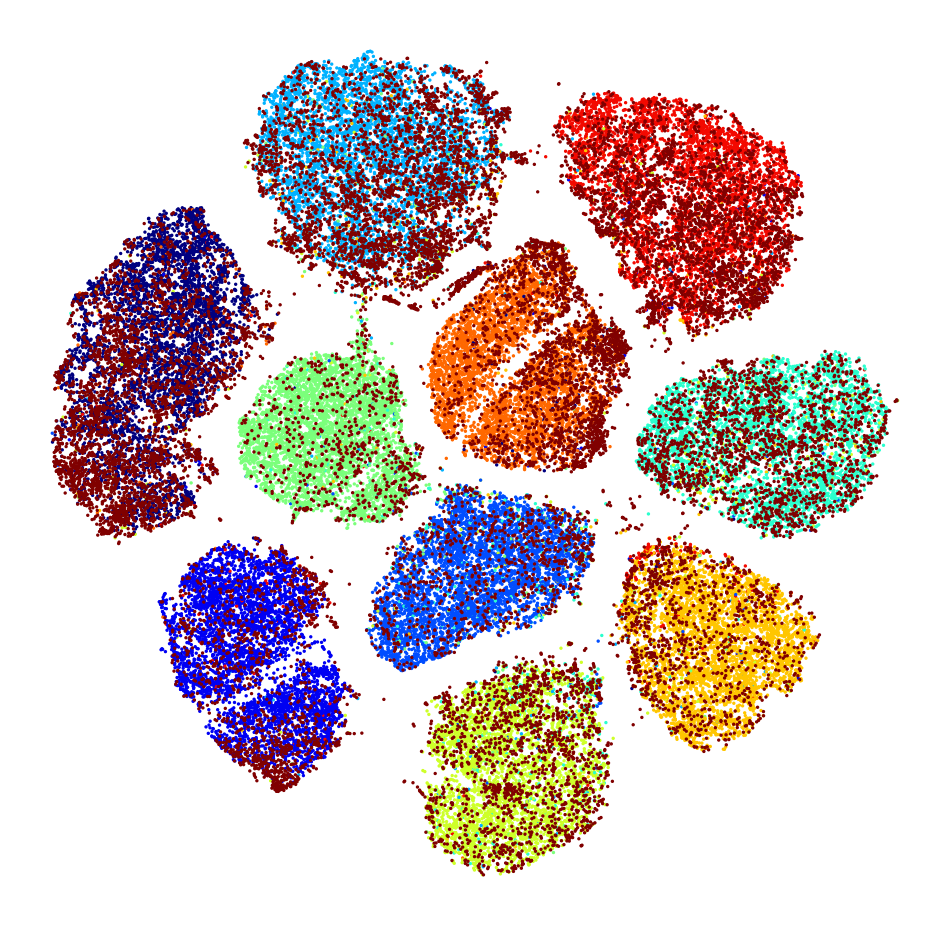}
        \caption{CIFAR-100 as OOD (DivideMix)} \label{fig:dm-c100}
    \end{subfigure}
    \begin{subfigure}[b]{0.33\textwidth}
        \centering
        \includegraphics[width=\linewidth]{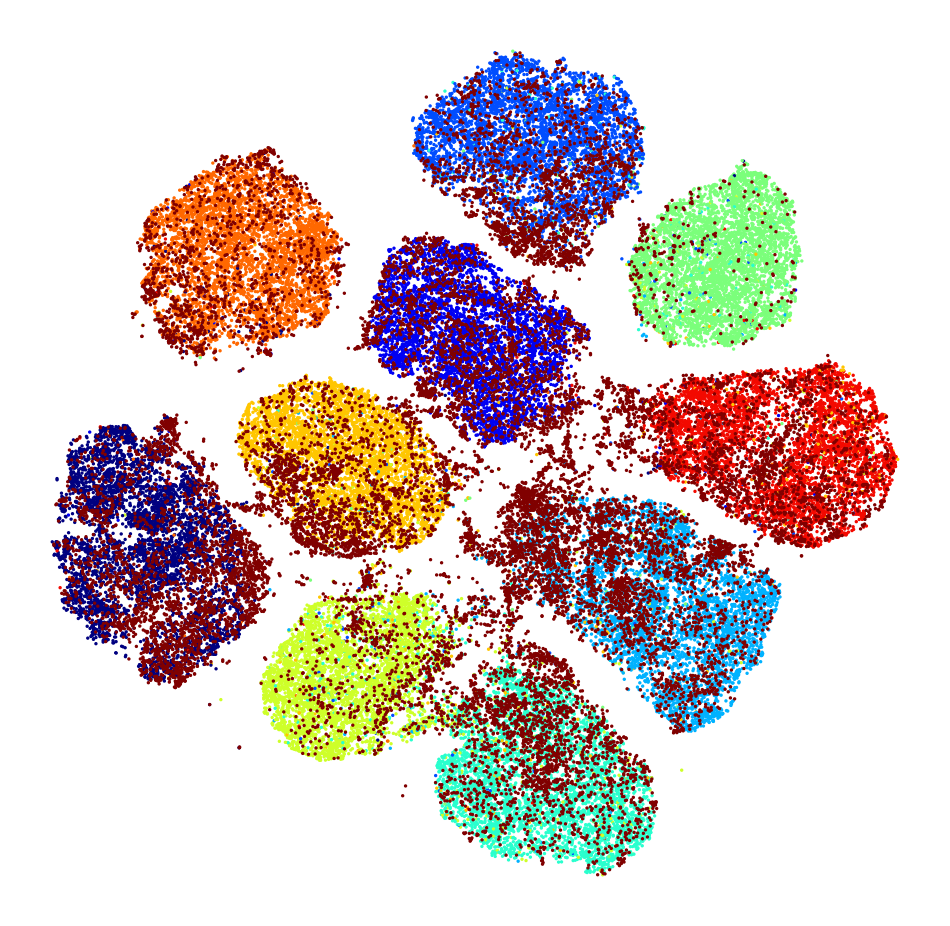}
        \caption{TinyImageNet as OOD (DivideMix)}
        \label{fig:dm-tin}
    \end{subfigure}
    \begin{subfigure}[b]{0.33\textwidth}
        \centering
        \includegraphics[width=\linewidth]{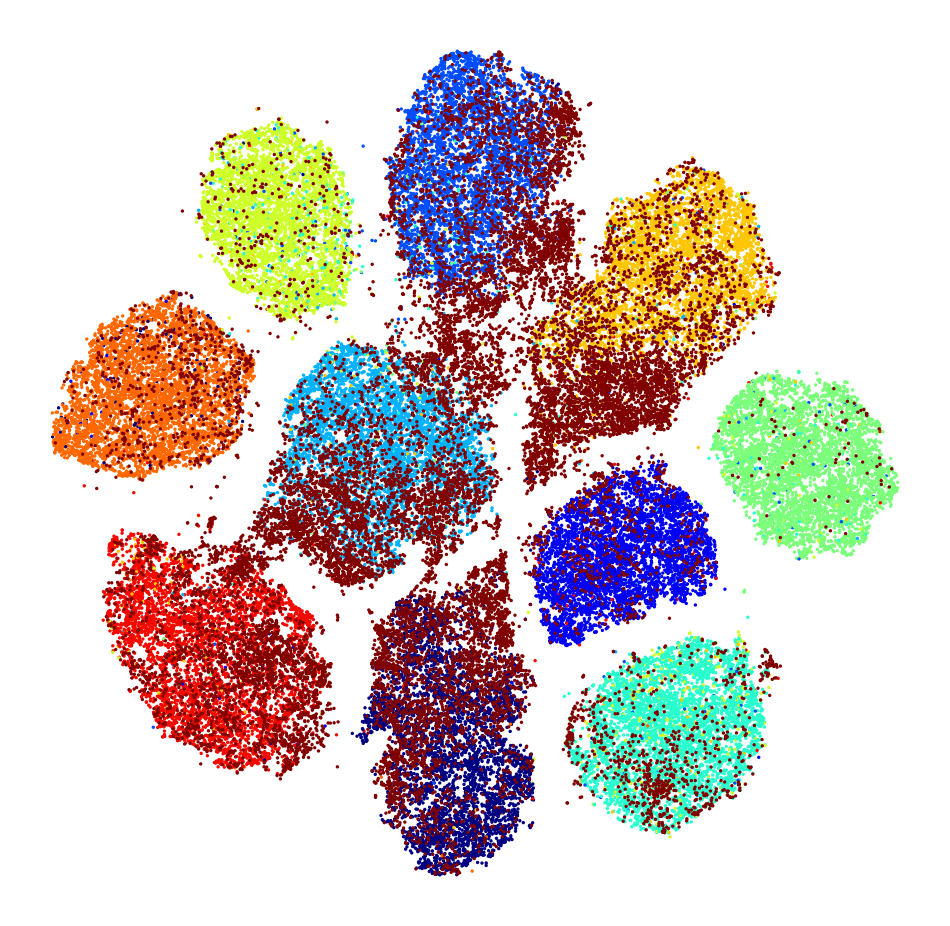} 
        \caption{Places-365 as OOD (DivideMix)}
        \label{fig:dm-places}
    \end{subfigure}
    
    \begin{subfigure}[b]{0.33\textwidth}
        \centering
        \includegraphics[width=\linewidth]{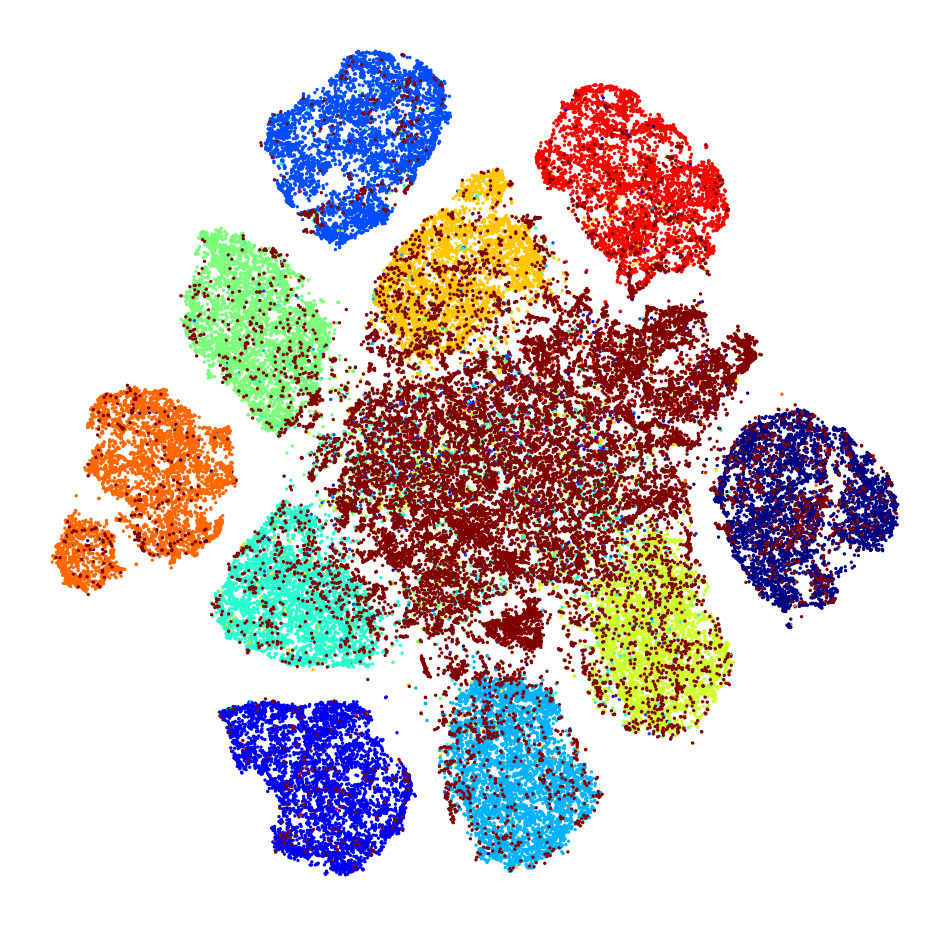}
        \caption{CIFAR-100 as OOD (Ours)} \label{fig:ngc-c100}
    \end{subfigure}
    \begin{subfigure}[b]{0.33\textwidth}
        \centering
        \includegraphics[width=\linewidth]{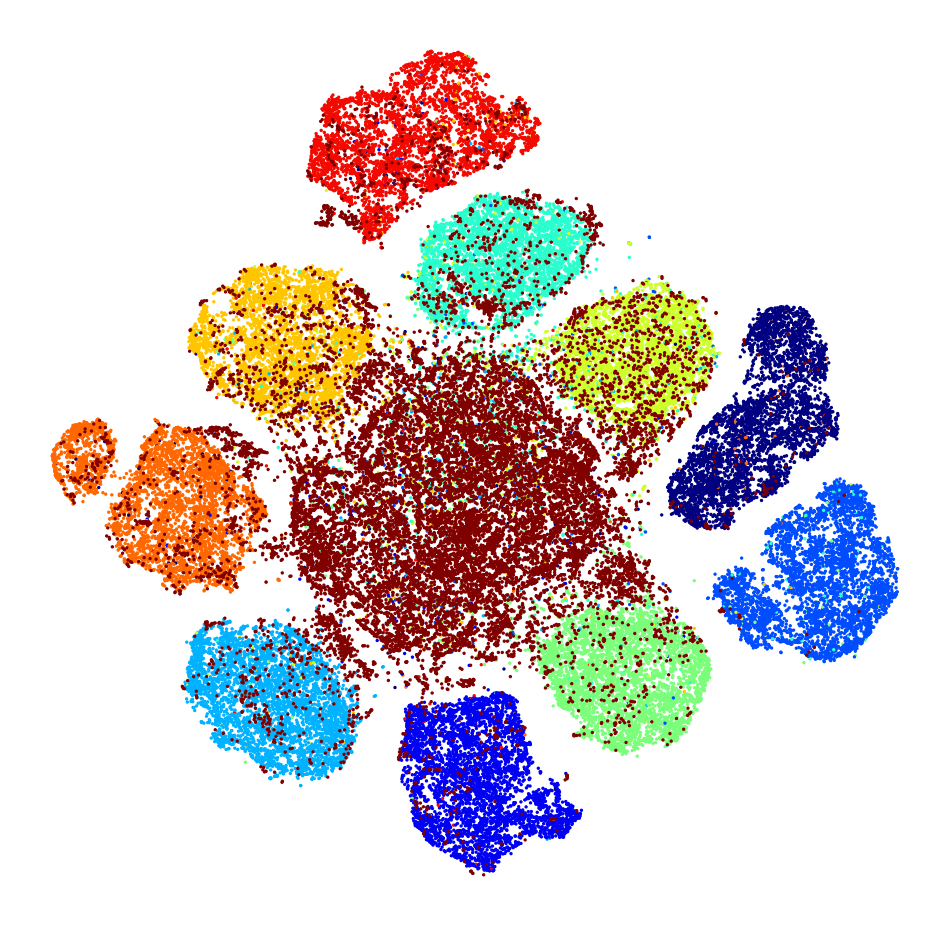}
        \caption{TinyImageNet as OOD (Ours)}
        \label{fig:ngc-tin}
    \end{subfigure}
    \begin{subfigure}[b]{0.33\textwidth}
        \centering
        \includegraphics[width=\linewidth]{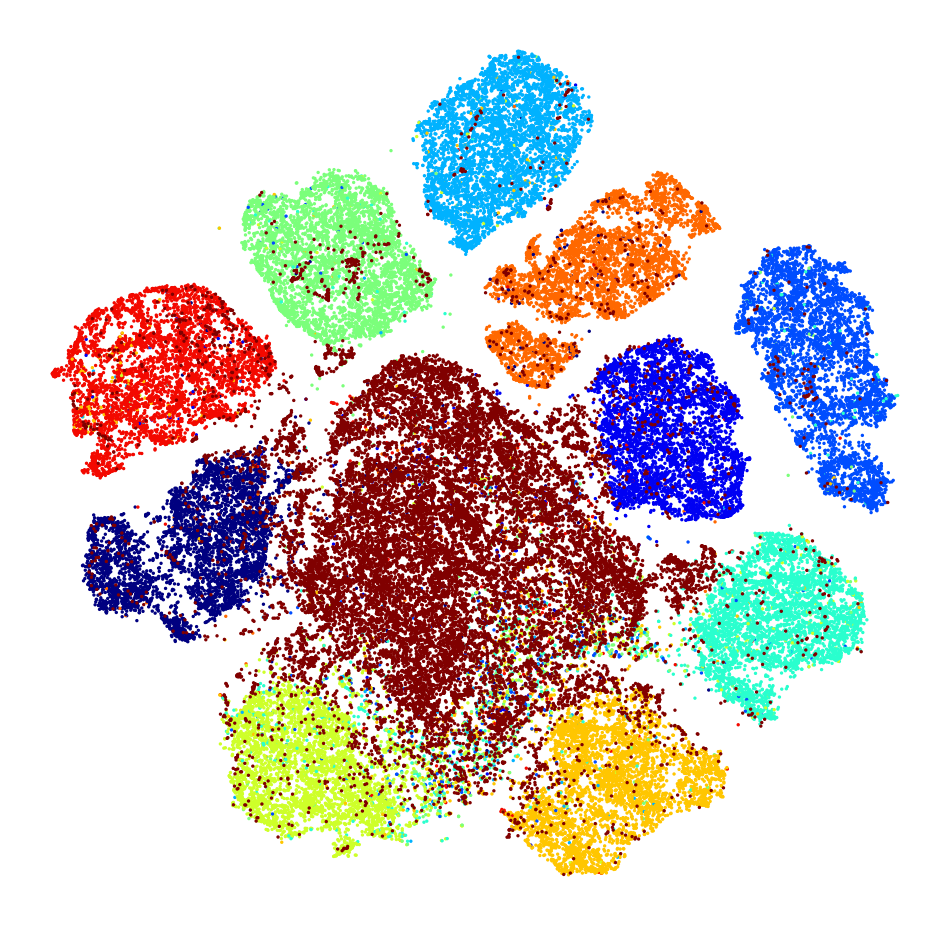} 
        \caption{Places-365 as OOD (Ours)}\label{fig:ngc-places}
    \end{subfigure}

    \caption{t-SNE visualization of learned feature representation. CIFAR-10 with 50\% sym. noise is used as IND dataset and 20k OOD samples are added for all experiments. The OOD samples are represented by brown points.}
    \label{fig:tsne}
\end{figure*}

\begin{figure*}[htb!]
    \centering
    \begin{subfigure}[b]{0.26\textwidth}
        \centering
        \includegraphics[width=\linewidth]{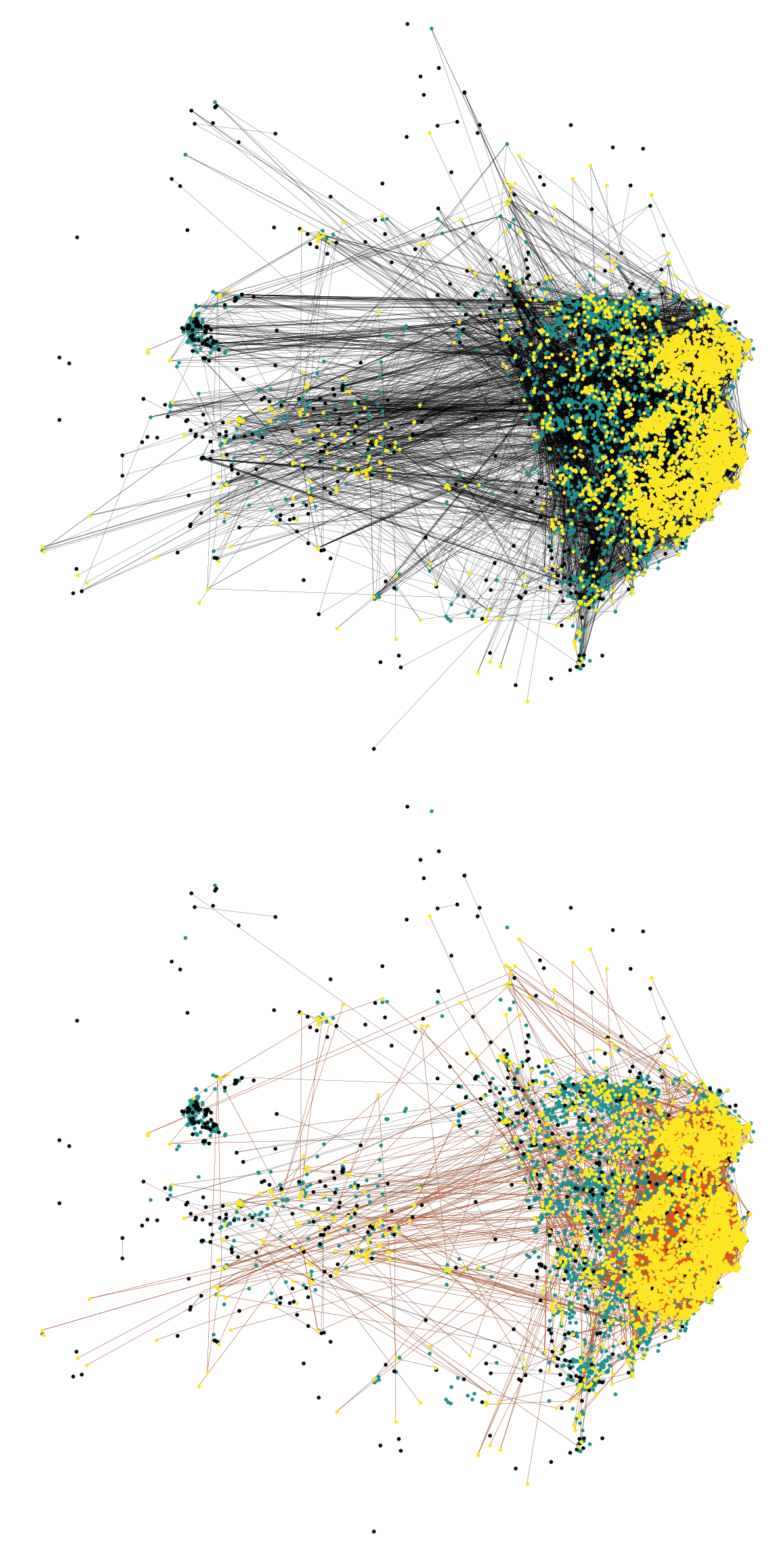}
        \caption{50 epoch} \label{fig:50e}
    \end{subfigure}
    \begin{subfigure}[b]{0.26\textwidth}
        \centering
        \includegraphics[width=\linewidth]{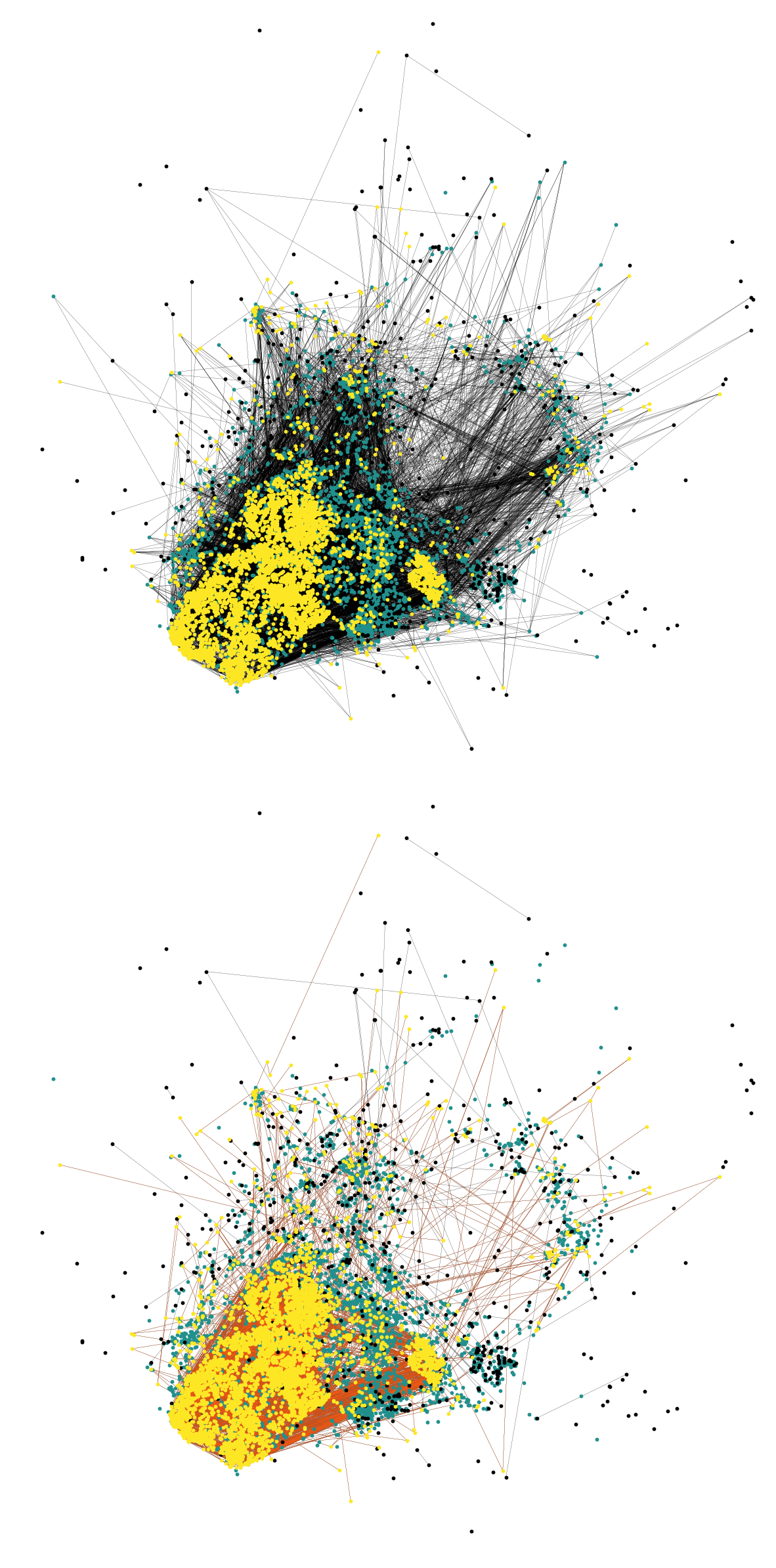}
        \caption{100 epoch}
        \label{fig:100e}
    \end{subfigure}
    \begin{subfigure}[b]{0.26\textwidth}
        \centering
        \includegraphics[width=\linewidth]{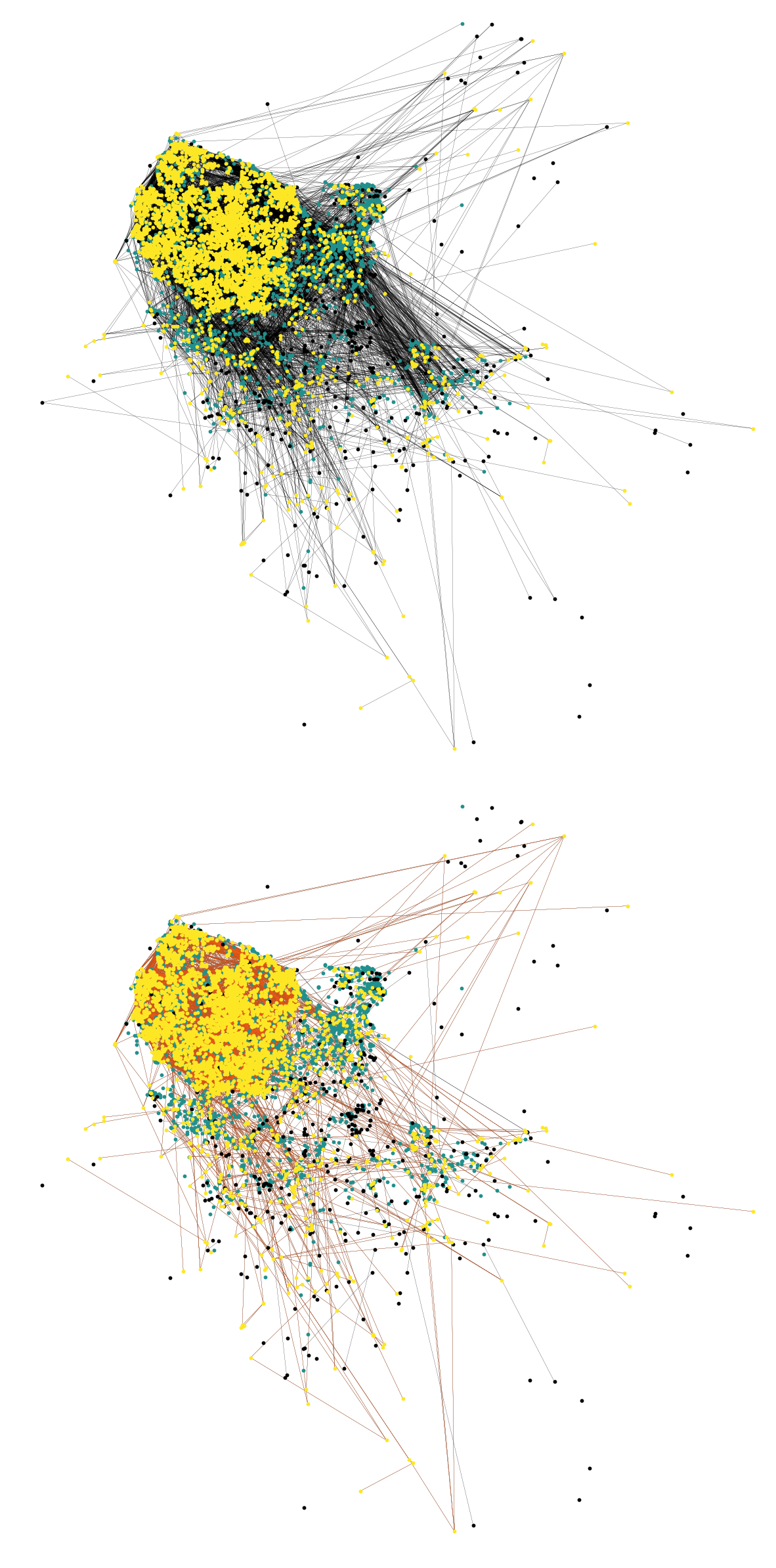} 
        \caption{150 epoch}
        \label{fig:150e}
    \end{subfigure}
    
    \begin{subfigure}[b]{0.26\textwidth}
        \centering
        \includegraphics[width=\linewidth]{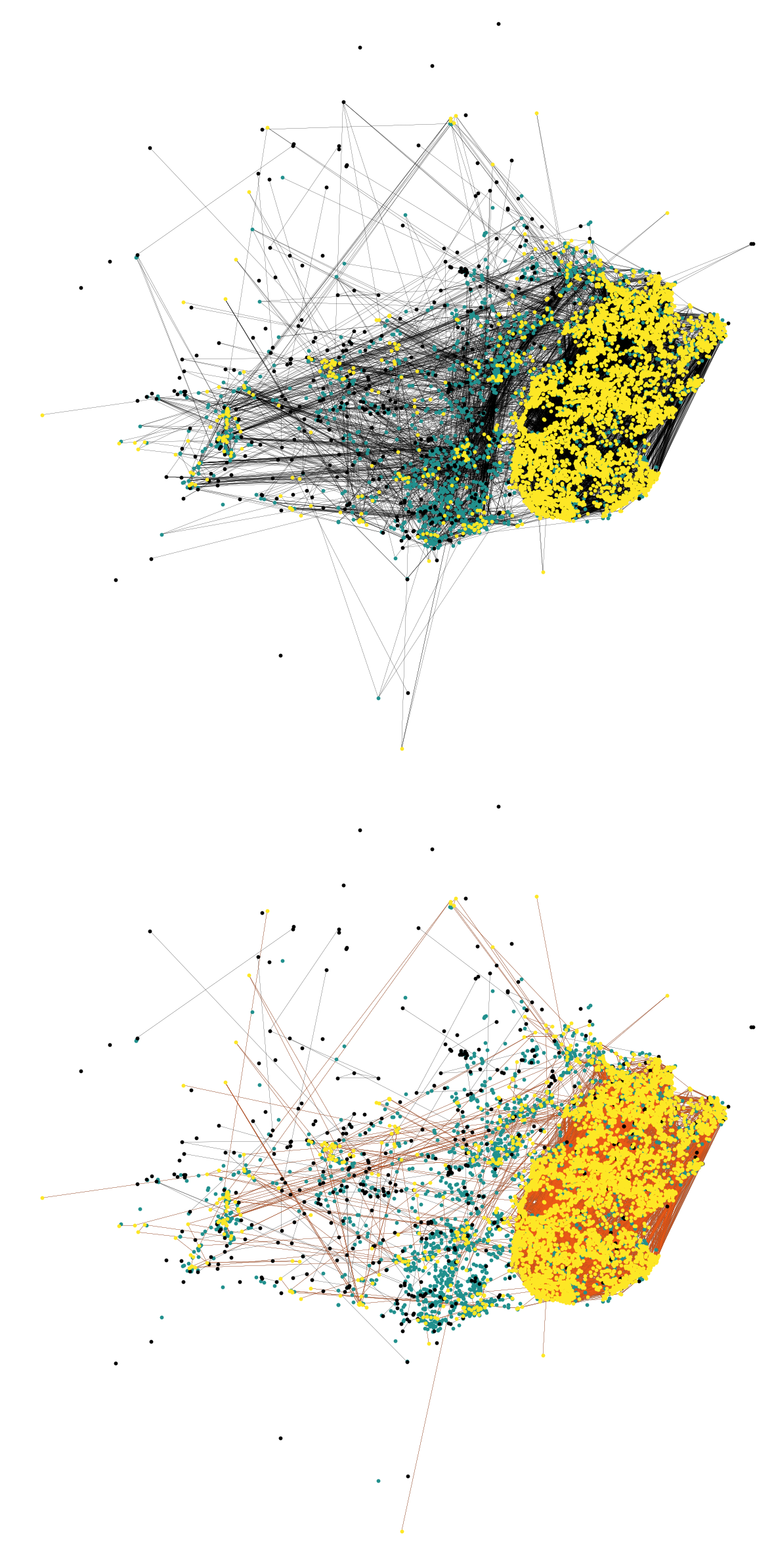}
        \caption{200 epoch} 
        \label{fig:200e}
    \end{subfigure}
    \begin{subfigure}[b]{0.26\textwidth}
        \centering
        \includegraphics[width=\linewidth]{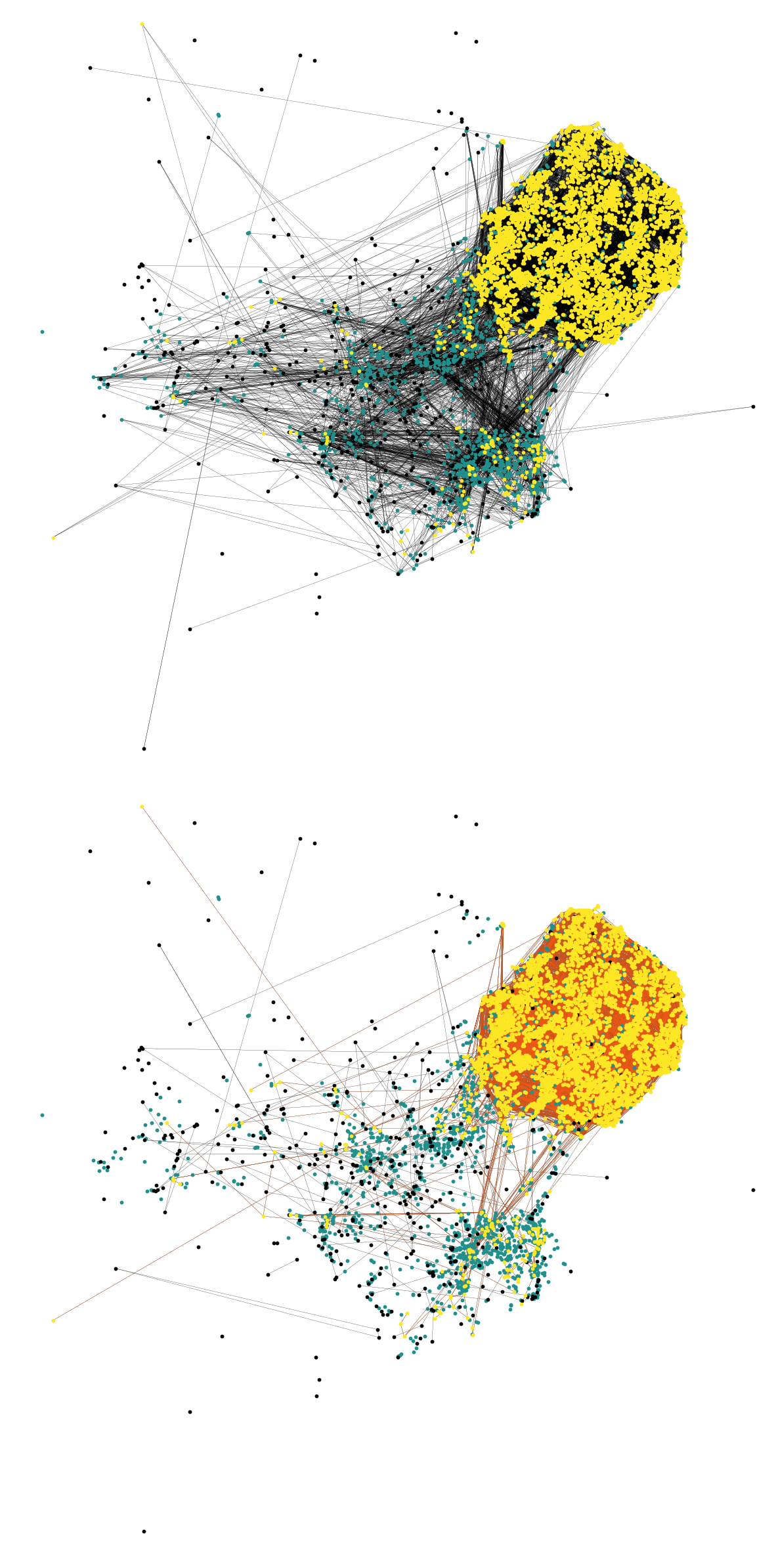}
        \caption{250 epoch}
        \label{fig:250e}
    \end{subfigure}
    \begin{subfigure}[b]{0.26\textwidth}
        \centering
        \includegraphics[width=\linewidth]{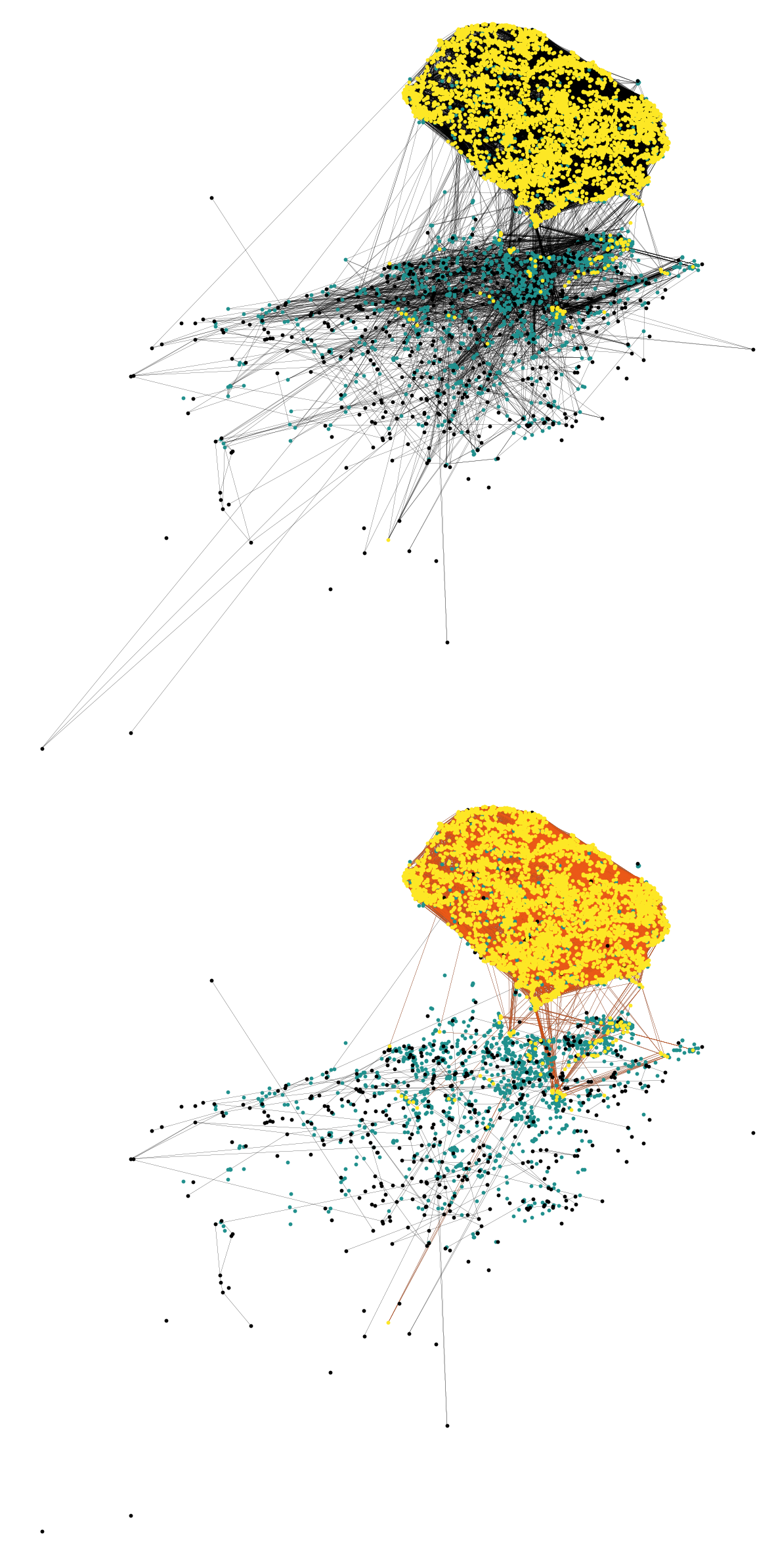} 
        \caption{300 epoch}
        \label{fig:300e}
    \end{subfigure}

    \caption{t-SNE visualization of the proposed subgraph selection at different training iterations. CIFAR-10 with 50\% sym. noise is used as IND dataset and 20k CIFAR-100 data are added as OOD samples. We draw all samples with pseudo-label 1. Green points represent samples removed by confidence-based selection and black points are samples removed by geometry-based selection. Points in yellow represent clean data selected by our method. Edges in the largest connected component are colored red. We visualize the constructed $k$-NN graph (top row) and the refined graph (bottom row) by performing our confidence-based selection.}
    \label{fig:tsne-edge}
\end{figure*}

\subsection{Visualization Results}
\textbf{Visualization of learned representation.} We visualize the learned feature representations of our method and DivideMix via t-SNE in Figure~\ref{fig:tsne}. CIFAR-10 with 50\% sym. noise is used as IND dataset and 20k OOD samples are added in each experiment. We use CIFAR-100, TinyImageNet, and Places-365 as OOD datasets for each experiment, respectively. The points in brown represent OOD samples, while samples with other colors are from CIFAR-10.~\Cref{fig:dm-c100,fig:dm-tin,fig:dm-places} show the learned representations of DivideMix, which are extracted from the last layer of the model. For comparison,~\Cref{fig:ngc-c100,fig:ngc-tin,fig:ngc-places} visualize the output of the projector $\operatorname{Proj}$ in our method. It can be observed that our method can learn more meaningful representations and separate OOD samples from IND samples effectively.

\textbf{Visualization of subgraph selection.} To further justify the efficacy of the proposed subgraph selection, we visualize the $k$-NN graph obtained at different training iterations in Figure~\ref{fig:tsne-edge}. CIFAR-10 with 50\% sym. noise is used as IND dataset and 20k CIFAR-100 data are added as OOD samples. We draw all the samples with pseudo-label 1. 
In these graphs, we use green points to represent samples removed by confidence-based selection while black points are samples removed by geometry-based selection. Points in yellow represent clean data selected by our method. The edges included in the largest connected component are in red. At different training iterations, we visualize the constructed $k$-NN graph (top row) and the refined graph (bottom row) by performing our confidence-based selection. As the training progresses, the feature representations of IND and OOD samples are gradually separated. Moreover, it can be seen that confidence-based selection significantly degrades the connectivity between clean samples and OOD samples, which can be further beneficial to geometry-based selection. As a consequence, samples retained by geometry-based selection distribute more and more compact in feature space. This observation justifies the validity of subgraph selection.

\subsection{Effectiveness of Noise Correction}
We demonstrate the effectiveness of graph-based noise correction on CIFAR-10 and CIFAR-100 datasets with 50\% symmetric noise. As shown in Figure~\ref{fig:lp}, As the training progresses, the noise rate continues decreasing. Our method reduces noise rate from 50\% to 4.24\% for CIFAR-10 and 14.67\% for CIFAR-100. This validates our noise correction methods can correct noisy labels effectively.

\subsection{Hyperparameter Sensitivity Analysis}

\textbf{Analysis of $\eta$ and $k$.}
We investigate the impact of $\eta$ for confidence-based selection and $k$ which is used to construct the $k$-NN graph. The results are shown in Figure~\ref{fig:eta} and~\ref{fig:k}. 
We vary $\eta$ from 0.6 to 0.9, and the test accuracy increases from 70\% to 72\%, showing that a small confidence threshold results in more label noise being included. AUROC increases from 72.08 to 92.23, this is because a higher threshold $\eta$ can filter out more OOD noisy samples, which can be further beneficial for representation learning and the calculation of prototypes. 
As for the parameter $k$, we choose its value from $\{50, 100, 150, 200\}$. It can be seen that~\algo~achieves similar performance with different values except $k=50$. The reason is that when $k$ is too small, the $k$-NN graph is very sparse, resulting in fewer data points being obtained from the largest connected component, hence only a few clean samples are selected for training. 

\textbf{Analysis of $\zeta$.} We report F-measure under best threshold $\zeta$ in Table~\ref{exp:ind-ood-detection-fmeasure-zeta}. Even with fixed $\zeta$ from 0.5 to 0.7, our method is robust enough and outperforms other methods with their best values of $\zeta$ in most cases. Here we report results for $\zeta=0.5$ and $\zeta=0.7$. We also report the standard deviation of best $\zeta$, which shows the stability of our method.

\begin{table}[h]
\setlength{\tabcolsep}{1pt}
\scriptsize
\centering
\caption{F-measure (threshold $\zeta$). IND dataset is with 50\% symmetric noise, 20k and 10k OOD samples are added into training set and test set, respectively. \textbf{Bold}: best; \uline{Underlined}: 2nd \& 3rd.} \label{exp:ind-ood-detection-fmeasure-zeta}
\begin{tabular}{ c | c | c c c c c c}
\toprule
IND & OOD & MSP & ODIN & MD & Ours & Ours$_{\zeta=\text{0.50}}$ & Ours$_{\zeta=\text{0.70}}$\\
\midrule
\multirow{4}{*}{\shortstack{C-10}} 
 & C-100 & 0.698$_{\text{(0.81)}}$ & 0.681$_{\text{(0.83)}}$ & 0.635$_{\text{(0.36)}}$ & \bf 0.838$_{\text{(0.55)}}$ & \uline{0.835}$_{\text{(0.50)}}$ & \uline{0.788}$_{\text{(0.70)}}$  \\
 \cmidrule{2-8}
 & TIN & 0.726$_{\text{(0.83)}}$ & 0.707$_{\text{(0.85)}}$ & 0.702$_{\text{(0.33)}}$ & \bf 0.875$_{\text{(0.54)}}$ & \uline{0.873}$_{\text{(0.50)}}$ & \uline{0.802}$_{\text{(0.70)}}$ \\
 \cmidrule{2-8}
 & P-365 & 0.717$_{\text{(0.81)}}$ & 0.705$_{\text{(0.14)}}$ & 0.651$_{\text{(0.40)}}$ & \bf 0.887$_{\text{(0.56)}}$ & \uline{0.882}$_{\text{(0.50)}}$ & \uline{0.827}$_{\text{(0.70)}}$ \\
\midrule
\multirow{2}{*}{\shortstack{C-100}} 
 & TIN & 0.687$_{\text{(0.41)}}$ & 0.705$_{\text{(0.02)}}$ & 0.526$_{\text{(0.38)}}$ & \bf 0.773$_{\text{(0.67)}}$ & \uline{0.743}$_{\text{(0.50)}}$ & \uline{0.770}$_{\text{(0.70)}}$  \\
 \cmidrule{2-8}
 & P-365 & 0.685$_{\text{(0.37)}}$ & \uline{0.696}$_{\text{(0.01)}}$ & 0.541$_{\text{(0.39)}}$ & \bf 0.731$_{\text{(0.70)}}$ & 0.687$_{\text{(0.50)}}$ & \uline{0.731}$_{\text{(0.70)}}$ \\
\midrule[0.7pt]
\multicolumn{2}{ c |}{${\zeta}$ (stand. dev.)} & ${ \text{0.21}}$ & ${ \text{0.39}}$ & ${\text{0.02}}$ & ${ \text{0.07}}$ & ${ \text{0.00}}$ & ${ \text{0.00}}$ \\
\bottomrule
\end{tabular}
\end{table}

\subsection{Performance of Model Ensemble}
Since model ensemble has shown to be useful when dealing with noisy data, we ensemble the outputs of two networks during testing phase and report the results in Table~\ref{exp:ensemble-results}. The complete DivideMix (DM) is used for comparison. Results show that our method outperforms DivideMix in most cases.

\subsection{Impact of AugMix on WebVision-50}

To better reveal the superiority of our method, we conduct ablation studies for AugMix on WebVision-50 dataset. The results are reported in Table~\ref{exp:webvision-augmix}. First, it can be seen that AugMix does help enhance the performance. Second, without applying AugMix, our method consistently outperforms strong baselines, i.e., ELR and DivideMix. The results further demonstrate the effectiveness of our method.

\setlength{\tabcolsep}{1.3pt}
\begin{table}[ht]
\small
\centering
\caption{Test accuracy (\%) using model ensemble. $^{+}$~indicates ensemble models.}\label{exp:ensemble-results}
\begin{tabular}{l|c c c c | c |c c c c}
\toprule
Data & \multicolumn{5}{ c |}{CIFAR-10} & \multicolumn{4}{ c }{CIFAR-100} \\
\midrule
Type & \multicolumn{4}{ c |}{Sym.} & Asym. & \multicolumn{4}{ c }{Sym.} \\
\midrule
Ratio & 20\% & 50\% & 80\% & 90\% & 40\% & 20\% & 50\% & 80\% & 90\% \\
\midrule
DM & 95.0 & 93.7 & \bf 92.4 & 74.2 & \bf 91.4 & 74.8 & 72.1 & 57.6 & 29.2 \\
Ours & \bf 95.88 & \bf 94.54 & 91.59 & \bf 80.46 & 90.55 & \bf 78.98 & \bf 75.91 & \bf 62.70 & \bf 29.76 \\
\midrule
DM$^{+}$ & 95.7 & 94.4 & \bf 92.9 & 75.4 & \bf 92.1 & 76.9 & 74.2 & 59.6 & 31.0\\
Ours$^{+}$ & \bf 96.27 & \bf 95.09 & 92.20 & \bf 83.75 & 91.70 & \bf 81.08 & \bf 77.16 & \bf 64.00 & \bf 34.18\\
\bottomrule
\end{tabular}
\end{table}

\setlength{\tabcolsep}{7.5pt}
\begin{table}[htbp]
\small
\centering
\caption{Ablation study for AugMix on WebVision-50. $^{+}$~indicates ensemble models.}\label{exp:webvision-augmix}
\begin{tabular}{l|c|c|c|c}
\toprule
\multirow{2}{*}{Method}
& \multicolumn{2}{ c |}{WebVision} & \multicolumn{2}{ c }{ILSVRC12} \\
\cmidrule{2-5}
 & top-1 & top-5 & top-1 & top-5 \\
\midrule
Ours (w/ AugMix) & 79.16 & 91.84 & 74.44 & 91.04 \\
\midrule
ELR & 76.26 & 91.26 & 68.71 & 87.84 \\
Ours (w/o AugMix) & \bf 77.56 & \bf 91.36 & \bf 72.92 & \bf 91.32 \\
\midrule
%\hline
DivideMix$^+$ & 77.32 & 91.64 & \bf 75.20 & 90.84 \\
ELR$^+$ & 77.78 & 91.68 & 70.29 & 89.76 \\
Ours$^+$ (w/o AugMix) & \bf 79.08 & \bf 91.80 & 75.12 & \bf 91.72 \\
\bottomrule
\end{tabular}
\end{table}

\subsection{Comparison with ProtoMix}
As one of the most recent related works, ProtoMix~\cite{li2021learning} employs unsupervised contrastive loss and mixup prototypical contrastive loss to learn robust representations, which can address different types of noisy data. We report the comparison results of NGC and ProtoMix on LOND task in Table~\ref{tab:acc-auroc-fmeasure}. For all experiments, we inject 50\% symmetric IND noise. 20k and 10k OOD samples are randomly selected and added into training set and test set, respectively. Although ProtoMix is not designed to detect OOD examples at test time, it is natural to achieve this by measuring the similarity between test examples and class prototypes, as shown in Eq. (9) in the main text. From the results, we can observe that NGC achieves better or comparable results in test accuracy. Regarding AUROC and F-measure, NGC consistently outperforms ProtoMix in all cases. Recall that, ProtoMix identifies IND and OOD noise according to predictive confidence, which means samples with high predictive confidence are determined as clean. As a result, many noisy samples are likely to be misidentified as DNNs gradually fit the training data. NGC overcomes this problem by exploiting the geometric structure of data. For each class, confident samples that clustered together are further selected by calculating the largest connected component. Our belief is that clean samples of the same class should distribute closely to each other, while noisy samples are pushed away. By first performing confidence-based selection, it breaks the connection between noisy and clean samples in the graph, which facilitates our geometry-based selection. Consequently, NGC excludes more noisy samples from training and achieves better performance.

\begin{table}[h]
\setlength{\tabcolsep}{4pt}
\small
\centering
\caption{Performance comparison of ProtoMix and NGC (Ours) on LOND task. 50\% symmetric IND noise is injected into training set, 20k and 10k OOD samples are added into training set and test set, respectively.} \label{tab:acc-auroc-fmeasure}
\begin{tabular}{ c | c | c c c }
\toprule
\multirow{2}{*}{\shortstack{IND}} & \multirow{2}{*}{\shortstack{OOD}} & Accuracy & AUROC & F-measure \\
\cmidrule{3-5}
& & \multicolumn{3}{ c }{ProtoMix / NGC} \\
\midrule
\multirow{3}{*}{\shortstack{C-10}}
 & C-100 & {\bf 92.51} / 92.31 & 84.64 / {\bf90.37} & 0.783 / {\bf0.838}  \\
 \cmidrule{2-5}
 & TIN & 93.12 / {\bf93.54} & 93.47 / {\bf94.18} & 0.862 / {\bf0.875} \\
 \cmidrule{2-5}
 & P-365 & 92.76 / {\bf93.67} & 94.14 / {\bf94.31} & 0.868 / {\bf0.887} \\
\midrule
\multirow{2}{*}{\shortstack{C-100}}
 & TIN & 72.80 / {\bf73.49} & 78.58 / {\bf94.24} & 0.653 / {\bf0.773} \\
 \cmidrule{2-5}
 & P-365 & 72.05 / {\bf73.44} & 75.19 / {\bf91.20} & 0.624 / {\bf0.731} \\
\bottomrule
\end{tabular}
\end{table}

\section{Pseudo-code of Our Proposed Method}
\begin{figure*}[!hbt]
    \centering
    \includegraphics[width=\linewidth]{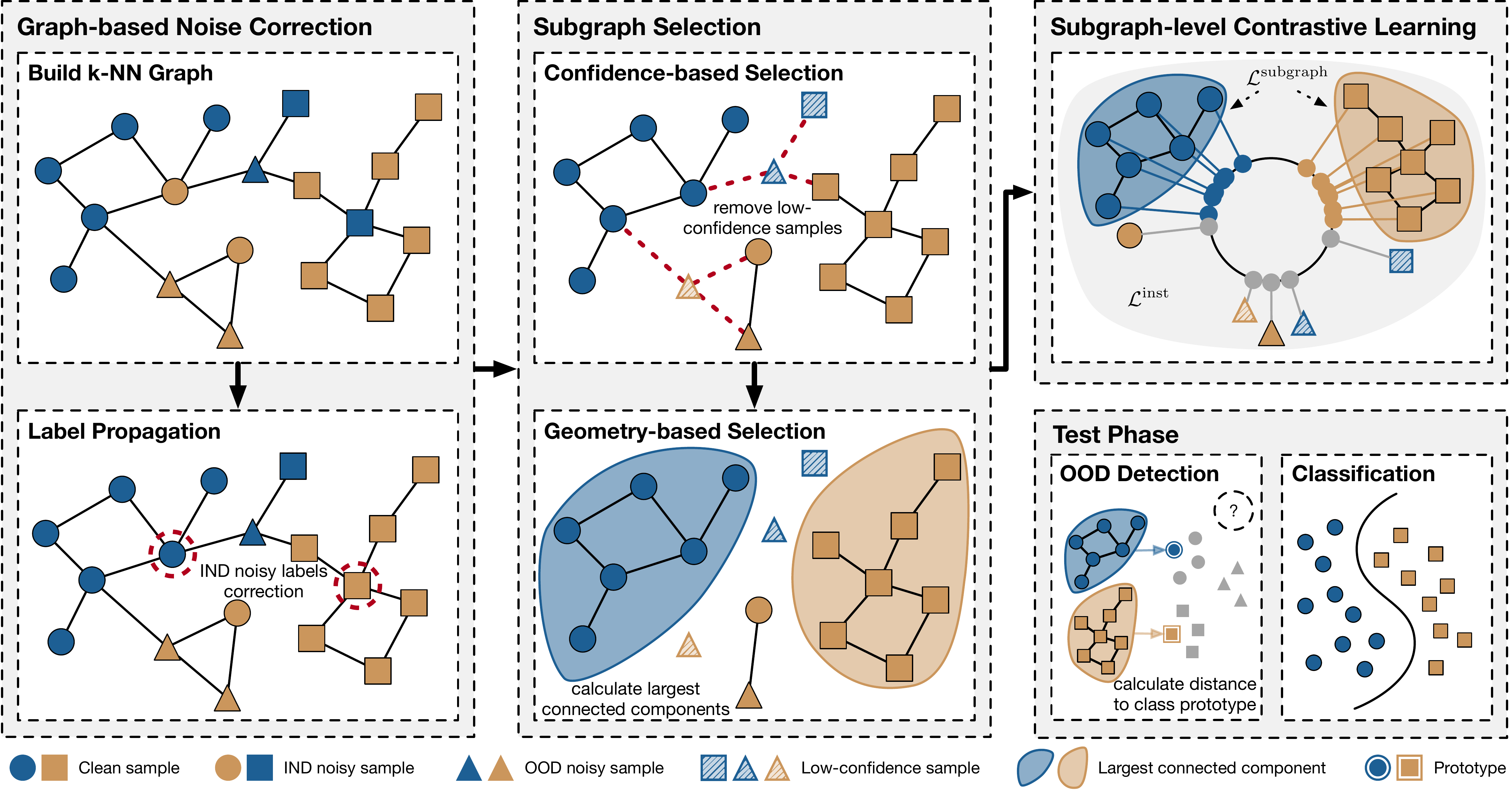}
    \caption{An illustration of proposed framework in binary classification case.}
    \label{fig:algorithm_illustration-full}
\end{figure*}

Algorithm \ref{alg:pseudo-code} lists the pseudo-code of NGC. For a better understanding of the proposed method, we illustrate the whole process in Figure \ref{fig:algorithm_illustration-full}.

\begin{algorithm*}[!hbt]
\caption{Noisy Graph Cleaning Procedure (one epoch)}
\begin{algorithmic}[1] %[1] enables line numbers
\STATE \textbf{Input:} training dataset $\{(\x_i, y_i)_{i=1}^N\}$, $k$-NN parameter $k$, confidence threshold $\eta$.
\STATE Construct the $k$-NN graph $G$ on training samples.
\STATE Refine soft pseudo-label $\tilde{\Y}_i$ for each sample $\x_i$ by performing graph-based noise correction on $G$.
\STATE If $\max_k \tilde{\Y}_{ik} < \eta$ and $\tilde{\Y}_{i y_i} \leq \frac{1}{K}$, remove the point $\x_i$ and its adjacent edges from the graph. 
\STATE The resulting graph is denoted by $\tilde{G}$.
\STATE Initialize the set of clean data $S = \emptyset$.
\FOR {$k = 1 \cdots K$}
\STATE Remove points that do not belong to class $k$ from graph $\tilde{G}$, i.e., $\hat{y}_i \neq k, \forall i \in [N]$. 
\STATE The resulting graph is denoted by $\tilde{G}(k)$.
\STATE Determine the connected components of $\tilde{G}(k)$ by disjoint-set data structures.
\STATE Remove small connected components of the graph $\tilde{G}(k)$, that is, only the largest connected component is retained. 
\STATE The resulting graph is denoted by $\tilde{G}(k)_{lcc}$. Points in $\tilde{G}(k)_{lcc}$ are treated as clean samples.
\STATE Update clean data set $S = S \cup \tilde{G}(k)_{lcc}$.
\ENDFOR
\STATE Calculate cross-entropy loss and subgraph-level contrastive loss on $S$.
\end{algorithmic}
\label{alg:pseudo-code}
\end{algorithm*}

\end{appendices}
\end{document}